\documentclass[letterpaper]{article} 
\usepackage{aaai25}  
\usepackage{times}  
\usepackage{helvet}  
\usepackage{courier}  
\usepackage[hyphens]{url}  
\usepackage{graphicx} 
\usepackage{natbib}  
\usepackage{caption} 
\usepackage{amsmath}	
\usepackage{amssymb}	
\usepackage{booktabs}
\usepackage{times}
\usepackage{microtype}
\usepackage{epsfig}
\usepackage{caption}
\usepackage{float}
\usepackage{placeins}
\usepackage{color, colortbl}
\usepackage{enumitem}
\usepackage{tabularx}
\usepackage{xstring}
\usepackage{multirow}
\usepackage{xspace}
\usepackage{url}
\usepackage{subcaption}
\usepackage{xcolor}
\usepackage[hang,flushmargin]{footmisc}
\pdfoutput=1
\usepackage{algorithm}
\usepackage{algorithmic}
\definecolor{mygray}{RGB}{242, 242, 242}
\definecolor{mycyan}{RGB}{244, 250, 252}
\usepackage{newfloat}
\usepackage{listings}
\DeclareCaptionStyle{ruled}{labelfont=normalfont,labelsep=colon,strut=off} 
\floatstyle{ruled}
\newfloat{listing}{tb}{lst}{}
\floatname{listing}{Listing}
\title{RemDet: Rethinking Efficient Model Design for UAV Object Detection} 
\author{
    Chen Li\textsuperscript{\rm 1,2},
    Rui Zhao\textsuperscript{\rm 1,2},
    Zeyu Wang\textsuperscript{\rm 1,2},
    Huiying Xu\textsuperscript{\rm 1,2}\thanks{Corresponding Author.},
    Xinzhong Zhu\textsuperscript{\rm 1,2}$^\ast$
}
\affiliations{
    \textsuperscript{\rm 1}School of Computer Science and Technology, Zhejiang Normal University, Zhejiang, 321004, China\\
    \textsuperscript{\rm 2}Research Institute of Hangzhou Artificial Intelligence, Zhejiang Normal University, Zhejiang, 311231, China \\

    \{lilsodachen,zhaorui,14797857499,xhy,zxz\}@zjnu.edu.cn
}

\usepackage{bibentry}

\begin{document}

\maketitle

\begin{abstract}
Object detection in Unmanned Aerial Vehicle (UAV) images has emerged as a focal area of research, which presents two significant challenges: i) objects are typically small and dense within vast images; ii) computational resource constraints render most models unsuitable for real-time deployment. Current real-time object detectors are not optimized for UAV images, and complex methods designed for small object detection often lack real-time capabilities. To address these challenges, we propose a novel detector, RemDet (Reparameter efficient multiplication Detector). Our contributions are as follows: 1) Rethinking the challenges of existing detectors for small and dense UAV images, and proposing information loss as a design guideline for efficient models. 2) We introduce the ChannelC2f module to enhance small object detection performance, demonstrating that high-dimensional representations can effectively mitigate information loss. 3) We design the GatedFFN module to provide not only strong performance but also low latency, effectively addressing the challenges of real-time detection. Our research reveals that GatedFFN, through the use of multiplication, is more cost-effective than feed-forward networks for high-dimensional representation. 4) We propose the CED module, which combines the advantages of ViT and CNN downsampling to effectively reduce information loss. It specifically enhances context information for small and dense objects. Extensive experiments on large UAV datasets, Visdrone and UAVDT, validate the real-time efficiency and superior performance of our methods. On the challenging UAV dataset VisDrone, our methods not only provided state-of-the-art results, improving detection by more than \textbf{3.4}\%, but also achieve \textbf{110} FPS on a single 4090. Code are available at \textit{https://github.com/HZAI-ZJNU/RemDet}.
\end{abstract}

%

\section{Introduction}
\label{sec:intro}
\begin{figure}[!tp]
    \centering
    \includegraphics[width=\linewidth]{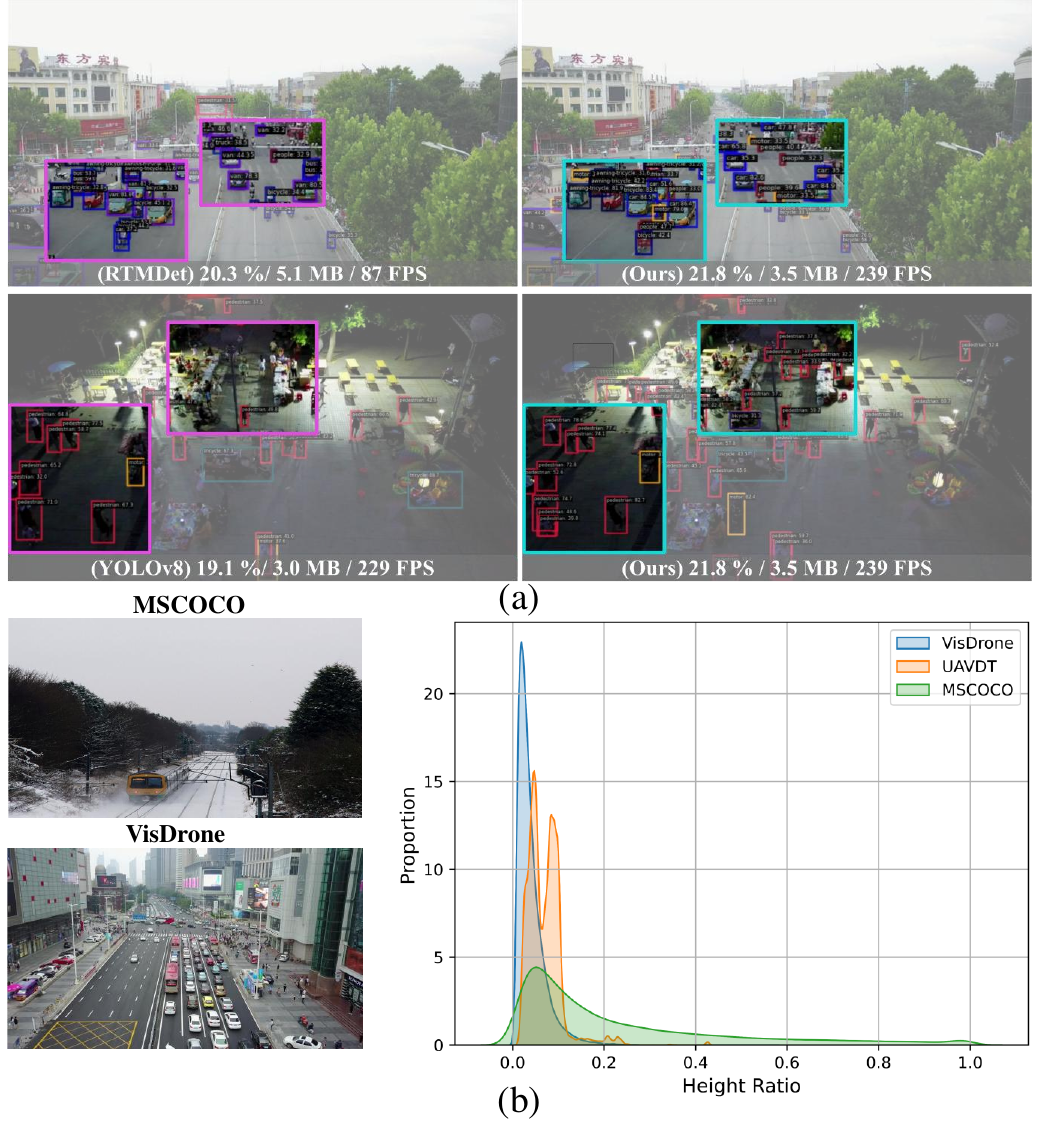}
    \caption{(a) Visualization of real-time detector results on VisDrone, white characters below the image represent mAP, model size, and FPS. (b) Kernel density analysis of detection object width on UAV and COCO datasets.}
    \label{fig:fig_visualize_datasets}
\end{figure}

Recent years have witnessed significant progress in object detection techniques, including the success of general detectors like Faster R-CNN \cite{fasterrcnn}, YOLO \cite{yolov1,yolov2}, and DETR \cite{detr}. Additionally, researchers have explored lightweight and efficient architectures tailored specifically for object detection. Despite these advancements, Unmanned Aerial Vehicle (UAV) images present unique challenges due to small and dense objects. Object detection in UAV images is a critical research area with applications in surveillance, disaster management, and environmental monitoring. Captured from an aerial perspective, UAV datasets exhibit a higher prevalence and density of small objects compared to traditional datasets (as illustrated in Figure \ref{fig:fig_visualize_datasets} (b)). For instance, while the MSCOCO \cite{mscoco} dataset contains an average of 7 objects per image, the VisDrone \cite{visdrone2019} dataset contains an average of 53 objects.
\begin{figure*}[!tp]
    \centering
    \includegraphics[width=\linewidth]{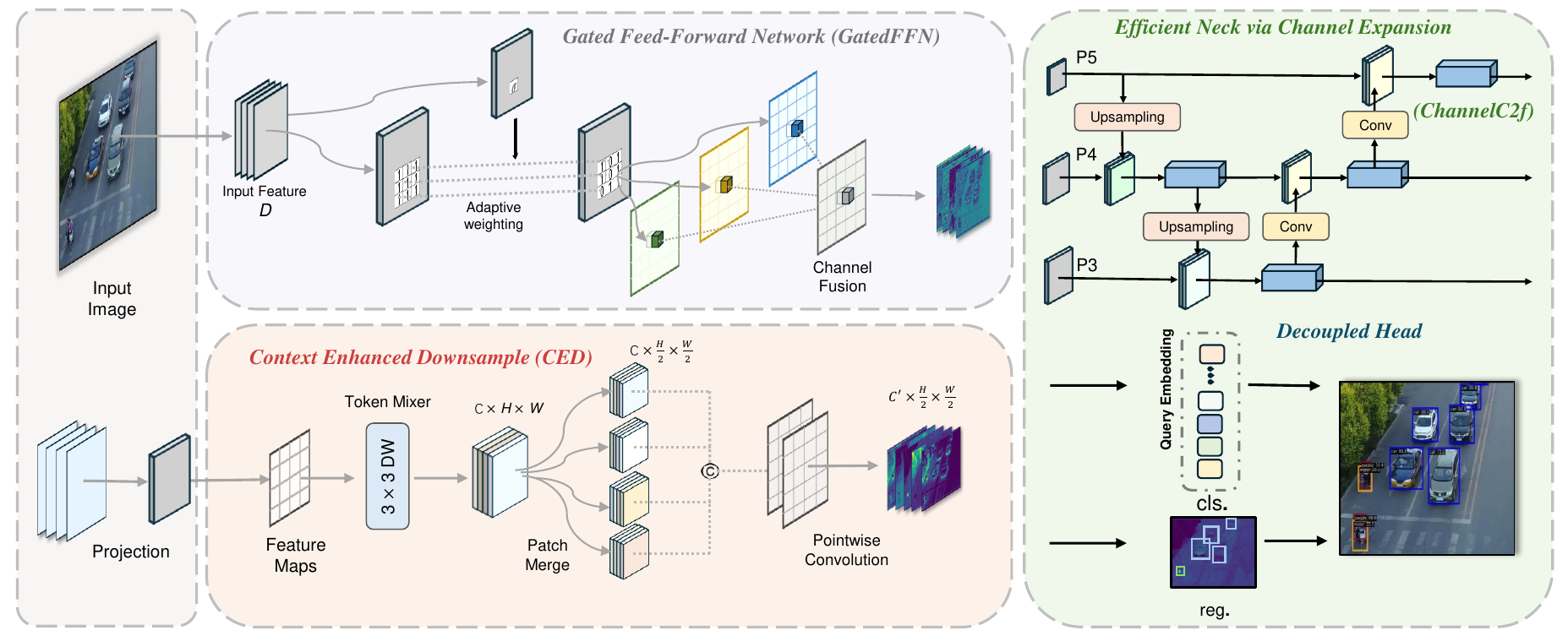}
    \caption{Overview of the proposed RemDet. Our method consists of three components: GatedFFN achieves adaptive weighting through multiplication, followed by channel fusion. CED utilizes patch merge operations to concatenate channels. ChannelC2f and decoupled heads are employed for prediction.}
    \label{fig:fig_architecture_remdet}
\end{figure*}

To address the detection challenges posed by small and dense objects, Region of Interest (RoI) methods are widely adopted.  \cite{roitransformer} and  \cite{dogfight} can amplify or prioritize the selection of regions of interest to enhance object visibility and distinguishability. Besides, current mainstream UAV detectors tend to employ density cropping methods \cite{czdet}, achieved through heavily handcrafted designs, to achieve background suppression or enhance fusion.

On the contrary, lightweight models have made significant progress in the field of generic object detection. By leveraging techniques such as depthwise separable convolutions \cite{mobilenetv3}, reparameterization \cite{ding2021repvgg}, and pruning \cite{pagcp}, they exhibit outstanding performance in object detection tasks. However, research specifically focused on lightweight models for UAV object detection remains relatively scarce. QueryDet \cite{querydet} and CEASC \cite{ceasc} use sparse convolutions in their detection heads to reduce model weights, which lowers computational requirements. However, they still rely on complex handcrafted designs and lack hardware optimization, hindering real-time efficiency.

This raises a question: Is it possible to balance the efficiency and accuracy of UAV detection through device-friendly operations rather than heavily handcrafted designs?

To address the challenges of detecting small and dense objects as well as achieving high-speed inference, in this paper, we propose RemDet (Reparameter efficient multiplication Detector), a one-stage anchor-free detector designed for real-time UAV object detection. Our approach rethinks the design of UAV detectors with the overarching goal of reducing information loss. Specifically, for small object detection, we introduce ChannelC2f and Context Enhanced Downsample (CED). The former extends the C2f with additional channels, providing a simple yet effective enhancement for small object detection. The latter combines the advantages of lightweight detectors \cite{mobilenetv2} and ViT \cite{vit} downsampling, effectively enhancing contextual information and reducing information loss. To meet the most demanding real-time detection requirements, we introduce GatedFFN. This model employs cost-efficient operations to achieve high-dimensional representation. Additionally, GatedFFN reparameterizes two convolutions, effectively balancing performance and speed. The maximum version,  RemDet-X, trained on high-resolution UAV images, achieves an mAP of 40\%, with inference latency as low as 9 ms on a single 4090, achieving 110 FPS.

The main contributions of our research include:
\begin{enumerate}
    \item We rethink the design of UAV detectors, discarding complex handcrafted designs. By exploring information loss, we effectively enhanced small object detection using a simplest structure.
    \item Following the principle of reducing information loss, our research revealed that high-dimensional representation alone can reduce information loss and enhance small object performance. We validate our analysis through empirical results (see Figure \ref{fig:fig_channel_impact_feature} (b)), theoretical exploration (in Section 3.2), and visual representations (Figure \ref{fig:fig_channel_impact_feature} (a)).
    \item To address the challenging real-time requirements, where complex designs and multi-feature fusion are impractical for accuracy improvement, our study reveals that multiplication, rather than feedforward networks, serves as a cost-effective and simpler high-dimensional representation. Our designs, based on this insight, reduces information loss while maintaining low latency. 
\end{enumerate}

\section{Related Work}
\label{sec:related}

\paragraph{Object Detection for UAV images} Unlike general object detection, UAV object detection has always focused on designing methods that transition from coarse to fine granularity. Addressing the non-uniform distribution of small objects in images, ClusDet \cite{yang2019clustered} utilized a clustering-based scale estimation method, effectively enhancing small object detection. UFPMP-Det \cite{ufpmpdet} first merged sub-regions provided by a coarse detector through clustering to suppress the background, then packaged the results into a mosaic for single inference. AMRNET \cite{amrnet} significantly expanded the coarse-to-fine framework through two specially designed modules. CZDet \cite{czdet}, based on density cropping method, first detected density-cropped regions and basic category objects during inference, then inputs them into the second stage of inference. Additionally, YOLC \cite{yolc} adaptively searched for clustered regions based on CenterNet \cite{centernet}, adjusted them to appropriate scales, and improves the loss function to enhance performance. However, most of these works are designed for detection heads or feature fusion layers, neglecting the information loss during the backbone stage. Above all, their real-time performance is also hindered by heavily handcrafted design.

\paragraph{Real-time detection of UAV images} In real-time UAV detection, one-stage detectors like YOLO \cite{yolov5,yolov6,yolov7wang2023} are widely used. The YOLO series has consistently aimed for real-time object detection, showcasing strong vitality through continuous updates and iterations. YOLOv8 \cite{yolov8}, in particular, improved real-time performance with its simple and effective C2f and decoupled head. However, on UAV images, the efficient extraction modules designed for these detectors often perform poorly due to background interference, as the objects to be detected are small and dense. Designing modules solely to enhance small object detection often fails to balance real-time performance.

Our work focuses on achieving a balance between small object detection and real-time performance, using more hardware-friendly designs rather than heavily handcrafted designs for UAV detection.

\section{Method}
\label{sec:method}
\subsection{Exploring Designs for Efficient Models}
\label{sec:3.1}
In order to further enhance model performance in complex scenarios, researchers have begun investigating factors that affect detection performance, with an increasing focus on the information bottleneck theory.

\paragraph{Principles for Hidden Layer Design} We rethink the information bottleneck definition to gain design insights. Simply put, the input variable is defined as $X$ and the output variable as $Y$. The hierarchical structure of a DNN forms a Markov chain, which can approximately represent all relationships and data. Each layer of the DNN relies solely on the input data from the previous layer, meaning that if any layer loses information about $Y$, it cannot be recovered in deeper layers.

\begin{figure}[!tp]
    \centering
    \includegraphics[width=\linewidth]{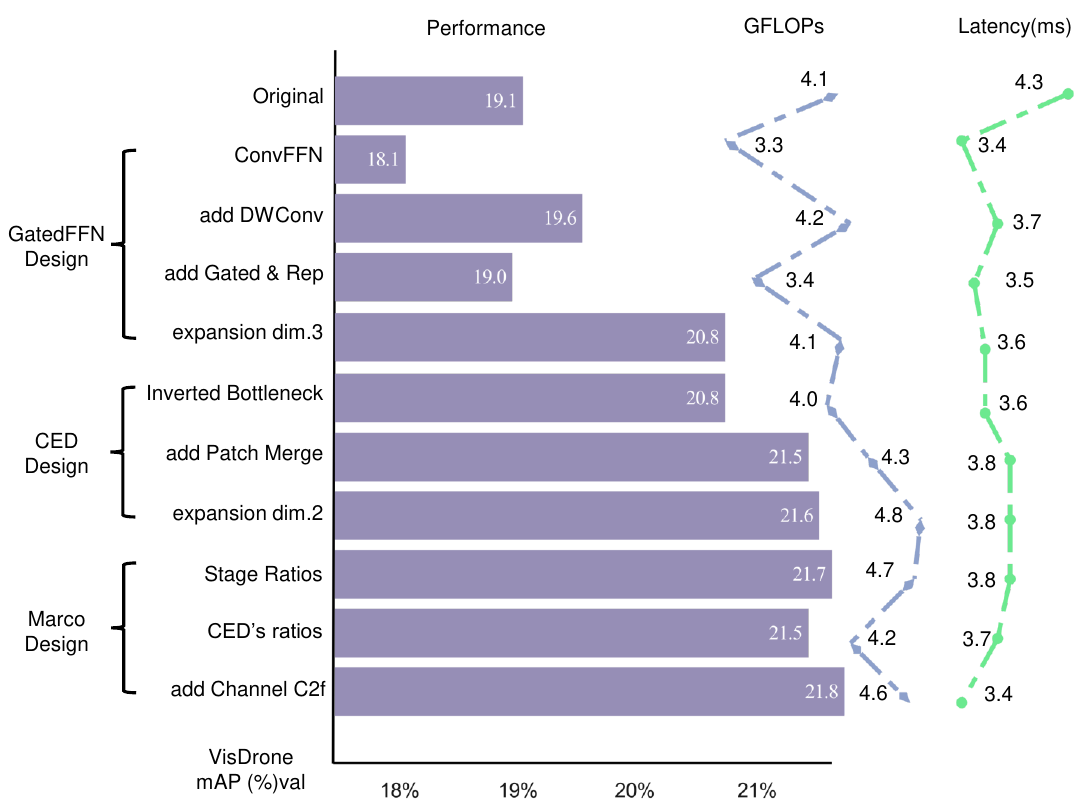}
    \caption{RemDet design process evaluated using FLOP, latency, and mAP.}
    \label{fig:fig_overall_design}
\end{figure}
We define the mutual information within the layer as $I(X;Y)$ and describe this process concisely using mathematics. The input variable $X$ has high resolution and low dimensionality, representing the lower-level representation of the data; whereas $Y$, as the prediction, has high dimensionality and low resolution. This implies that the neural network essentially performs data compression throughout the process. In statistics, we denote the X related to the prediction of $Y$ as $X^{\prime}$. Under the constraint $I(X^{\prime};Y)$, finding $X^{\prime}$ can be expressed as the minimization of the Lagrangian function. $I(X;Y|X^{\prime})$ represents the information between $X$ and $Y$ that is not captured by $X^{\prime}$, and by replacing the constraint with it, the formula can be equivalently expressed as:
\begin{align}
    L_{p(x^{\prime}|x)} = & I(X;X^{\prime}) + \beta I(X;Y|X^{\prime}) \label{eq1:lagrangian}
\end{align}
Where $\beta \geq 0$. We further explain its optimization objective: $\beta$ represents the relaxation variable balancing complexity $I(X;X^{\prime})$ and irrelevance $I(X;Y|X^{\prime})$. It can be observed that when irrelevance is 0 and $I(X;X^{\prime})$ is minimized, the Lagrangian function achieves its minimum value. This implies that $X^{\prime}$ also reaches its minimum value. In DNN, this indicates that $X^{\prime}$ should be as simple as possible, i.e., \textbf{finding the minimal information from $X$ that satisfies the conditions.}

Due to the complexity of the data $X$, it is difficult to find the minimal sufficient statistic. When using $Y^{\prime}$ for prediction, we have $I(X;X^{\prime}) \ge I(Y;Y^{\prime})$. More generally, for any neural network, it can be expressed as:
\begin{align}
    I(Y;X) \ge & I(Y;h_i) \ge I(Y;h_{i+1}) \ge I(Y;Y^{\prime})
\end{align}
Where $h_i$ represents the intermediate information. The above equation holds with equality only if each layer is a sufficient statistic of its input. Therefore, the goal of each layer is to optimally \textbf{capture all information relevant to the output in its input and discard all irrelevant parts.}
\begin{figure*}[!tp]
    \centering
    \includegraphics[width=\linewidth]{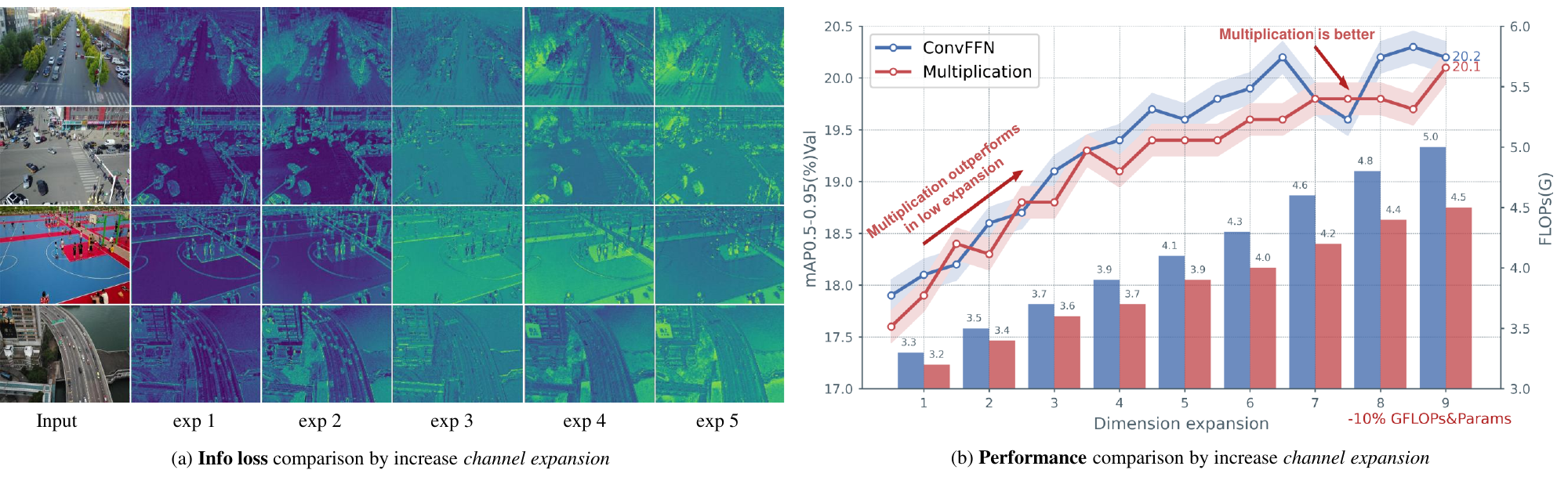}
    \caption{(a) Visualization of model features as channel expansion increases. (b) Comparison of Multiplication and ConvFFN.}
    \label{fig:fig_channel_impact_feature}
\end{figure*}
\paragraph{Dimension Expansion Design Principle} In \cite{tishby2015deep}, the importance of designing compact DNN is emphasized, focusing on reducing the number of layers and minimizing the units per layer. YOLOv9 \cite{yolov9} employs a feature-sharing approach to enable each layer to acquire relevant information from the preceding layer. While theoretically effective, this method incurs high training costs, posing significant challenges to model training. Conversely, other efficient structures in \cite{leyolo}, guided by the information bottleneck principle, utilize a simple inverted bottleneck structure to enhance inter-layer information interaction. However, this design lacks detailed explanation. In \cite{rethinkingchannel}, a search on channel dimensions was conducted, resulting in the design principle of \textbf{maintaining constant channel dimensions within layers and linearly increasing them between stages.}

In summary, the information bottleneck theory guides us in enhancing the mapping relationship between $X$ and $Y$. The core objective is to create a cleaner and more compact structure to \textbf{achieve simpler intra-layer mappings}. Additionally, the design between layers must maintain consistent input and output dimensions. This brings us to the next design goal: \textbf{enhancing intra-layer information}.

\subsection{How to Design Efficient Modules?} \label{sec:3.2}
\paragraph{Design for Enhanced Information Interaction}
After establishing inter-layer design guidelines, our focus shifts to enhance intra-layer information interaction. \cite{shwartz2017opening} delve into the essence of hidden layers, emphasizing their role in learning from input $X$ while compressing $Y$ information without labels. Our experimental investigations center on a Multilayer Perceptron (MLP), a common choice for dimension expansion when learning from $X$.

To assess the impact of this operation on UAV detection, we introduce ConvFFN, which employs only two 1$\times$1 convolutions as its backbone, with scaled hidden layer dimensions. The detailed design of ConvFFN is provided in the appendix. Surprisingly, with a channel expansion set to 3, ConvFFN performs comparably to a baseline \cite{yolov8} that includes dense computations and residual connections, while reducing parameters and computation by approximately 10\%. This phenomenon underscores the heightened optimization benefits of minimizing information loss in UAV datasets, where smaller objects prevail compared to general datasets. Visualizing features across different channel expansions (Figure \ref{fig:fig_channel_impact_feature} (a)), we observe a significant amplification in feature weights as hidden dimensions increase. This amplification is indicative of enhanced modeling capabilities. Under consistent input-output dimensions, the model more effectively accomplishes the tasks of ‘learning’ and ‘compression’. We deem that higher representations enhance inter-layer interaction and are a form of “learning”.

\paragraph{Multiplication Resulting in Higher Representations}
In deep learning, the MLP serves as a simple and common modeling approach, often used for processing input-output data. However, for pixel-sparse images with small correlations, the operations of MLP can become redundant. Therefore, we need a more efficient method for mapping to high-dimensional.

Gated Linear Units (GLU) \cite{glu} in natural language processing has been considered an alternative to recurrent neural networks (RNNs). We specifically focus on the gating part within GLU, which involves element-wise multiplication.

We analyze the dimension expansion of MLP and GLU from a mathematical perspective. To simplify the analysis, we just consider the case of one-output channel transformation, single-element input. To align the two methods, we assume that the input element $x \in \mathbb{R}^{d \times 1}$ and map it to $w_1^Tx$ and $w_2^Tx$. These two elements are then combined in an MLP. Specifically, the representation of MLP is
\begin{align}
     w^T_0x = & w_1^Tx+w_2^Tx \nonumber \\
     = & (\sum\limits_{i=1}^d w_1^ix^i ) + (\sum\limits_{j=1}^d w_2^jx^j )
     \label{eq_3}
\end{align}
Where $w_0\in \mathbb{R}^{d\times2}$. Represent the multiplication in the same way:
\begin{align}
   & w_1^Tx \ast w_2^Tx \nonumber\\
    = & (\sum\limits_{i=1}^d w_1^ix^i ) \ast (\sum\limits_{j=1}^d w_2^jx^j )\nonumber \\
    = & \sum\limits_{i=1}^{d} \sum\limits_{j=1}^{d} w_1^iw_2^jx^ix^j  \label{eq_4}
\end{align}
where $w_1,w_2\in \mathbb{R}^{d\times 1}$. By computing the polynomial sum, it obtain $\frac{(d+1)d}{2}$ distinct terms. From a parameter perspective, multiplication incurs no additional computational cost. Moreover, given that $d \gg 2$, we observe that $\frac{d^2 + d}{2} \geq 2d$, indicating higher dimensionality after element-wise multiplication. However, during multiplication, $w_2^Tx$ is discarded but implicitly included in the output. As a result, the dimension of the output is halved compared to the original. As the dimension increases, the impact of $w_2^Tx$ becomes more pronounced. This shows that when hidden layers are expanded in low dimensions, The closer the performance of multiplication will be to MLP. We conducted a series of experiments, as shown in Figure \ref{fig:fig_channel_impact_feature} (b). The results not only support our conjecture but also reveal that as the dimensionality extension increases, the gains in accuracy from higher dimensions for both methods become closer to each other. Further, the computational demand for multiplication is lower, allowing us to compensate for the implicit dimension loss resulting from polynomial addition by increasing dimension expansion. Notably, Figure \ref{fig:fig_channel_impact_feature} (b) illustrates that the computational cost of channel-expanded multiplication (to 9) is comparable to that of MLP (with 7), yet the mAP improves by 0.3\%. 

We also observe that the multiplication resembles the form of a kernel function. The kernel function is defined as $K(x, z) = \phi(x) \cdot \phi(z)$, where $\cdot$ is expressed as an inner product. In fact, we perform element-wise polynomial multiplication in formula \ref{eq_4}, with $w_2^Tx$ serving as the mapping function ($\phi(z)$) to increase the result’s dimensionality.

Consequently, \textbf{we now adopt the multiplication as our primary design approach.}

\begin{figure}[!tp]
    \centering
    \includegraphics[width=\linewidth]{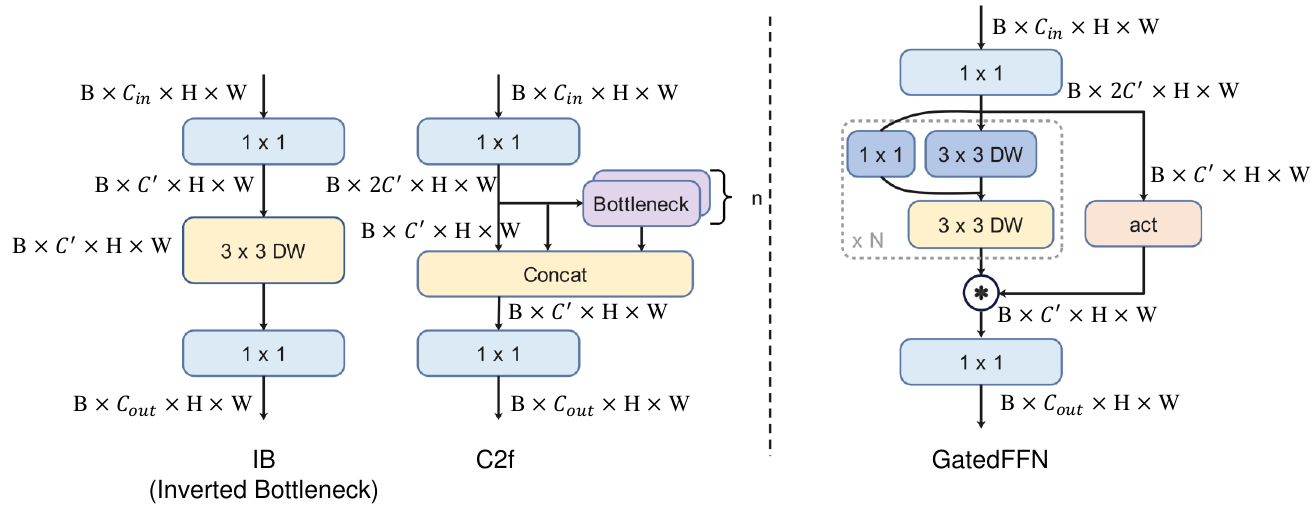}
    \caption{Structure of the IB, C2f, and proposed GatedFFN.}
    \label{fig:fig_gatedffn_design}
\end{figure}

\paragraph{Module Design} Our model design, as depicted in Figure \ref{fig:fig_gatedffn_design}, provides a clear exposition of our approach. Based on C2f, we employ a double-branch multiplication, eliminating the Bottleneck structure, and utilize 1$\times$1 and 3$\times$3 depthwise convolutions as reparameterized convolutions in the main branch. Notably, \cite{wang2020eca} emphasizes that direct channel compression may compromise expressive capacity. To address this concern, we set the channel expansion factor to 3, enhancing inter-layer information. Finally, a 1$\times$1 convolutional layer is positioned at the end of the model to compress in-layer information for efficient output. Collectively, these design choices constitute what we refer to as a lightweight structure—the GatedFFN, see Figure \ref{fig:fig_gatedffn_design}.

In the Neck layer, we merely scale the C2f (see in Figure \ref{fig:fig_gatedffn_design}) channels creating a structure known as ChannelC2f. Specifically, we increase the overall channel expansion from 0.5 to 1.0 and reduce the Bottleneck’s expansion ratio from 1 to 0.25, thereby minimizing dense computations. Thus, we enhance intra-layer information solely by adjusting channel expansions.

\subsection{Context Enhanced Downsample Module}
In neural networks, downsampling modules are used to reduce the resolution of feature maps. As far as we know, deepening the downsampling module is an efficient way to mitigate information loss resulting from resolution reduction. For this purpose, EfficientViT \cite{efficientvit} and RepViT \cite{repvitwang2024} deepen the module and incorporate an additional FFN at the end for information compression. In contrast, lightweight CNNs employ simpler modules, such as 3$\times$3 convolutions with a stride of 2 for downsampling, which is extremely fast. However, due to insufficient network depth, concerns arise regarding information loss and performance degradation. In ViT \cite{vit}, downsampling is typically achieved using Patch Merge layers, effectively increasing the channel expansion of layers to avoid information loss. Additionally, Convnext \cite{convnext} explores how to adapt CNNs using ViT designs in detail. However, due to the difficulty of Patch Merge in gaining an advantage over convolutions in classification tasks, exploration of downsampling layers was abandoned. This raises a question: Can combining depthwise separable convolutions and Patch Merge improve performance on UAV images?

To address this issue, we adopted the Inverted Bottleneck and changed the stride to 1 while setting the input dimension expansion to 1. It is worth noting that the performance of depthwise separable convolutions is limited by the loss of channel information during the split convolution operation. To enhance information capture, we inserted Patch Merge after the depthwise convolution, allowing the subsequent pointwise convolution to obtain richer information. Detailed design specifications can be found in the appendix. We named the entire module the Context Enhanced Downsample (CED). The CED module effectively mitigates information loss while maintaining high-speed inference.
\begin{table*}[!tp]
	\centering	
\tabcolsep=0.35cm
\resizebox{1\textwidth}{!}{
	\begin{tabular}[c]{lcc|ccc|ccc|cc}
		\toprule
		Model & imgsz  & test &  {AP$^{val}_{95}$} & {AP$^{val}_{50}$} & {AP$^{val}_{75}$} & {AP$^{val}_{s}$}  & {AP$^{val}_{m}$}  & {AP$^{val}_{l}$} & {Latency(ms)} & FLOPs(G)  \\
		\midrule
        YOLOv6-v3.0-N & 640 & o & 19.0 & 32.8 & 18.7 & 9.9 & 29.0 & 41.3 & 11.9 & 3.9 \\
        YOLOv8-N  & 640 & o & 19.1 & 33.0 & 18.9 & 10.6 & 28.9 & 38.3 & 4.3 & 4.1 \\
        YOLOv7-Tiny & 640  & o & 19.4 & 35.1 & 18.5 & 10.5 & 29.1 & 41.0 & 3.8 & 4.2 \\
        RTMDet-Tiny & 640  & o & 20.3 & 33.5 & 21.2 & 10.2 & 32.9 & \textbf{47.1} & 13.2 & 5.1 \\
        \rowcolor{mycyan}
        {\bf RemDet-Tiny} & 640  & o & \textbf{21.8} & \textbf{37.1} & \textbf{21.9} & \textbf{12.7} & \textbf{33.0} & 44.5 & 3.4 & 4.6 \\
        \midrule
        QueryDet  & 800  & o & 19.6 & 35.7 & 19.0 & - & - & - & 288 & -  \\
        RetinaNet  & 800  & o & 20.2 & 36.9 & 19.5 & - & - &- & 14.7 & 210 \\ 
        Faster-RCNN & 800  & o & 21.4 & 40.7 & 19.9 & 11.7 & 33.9 & 54.7 & 21.2 & 285 \\
        RTMDet-L & 640  & o & 23.7 & 37.4 & 25.5 & 12.5 & 38.7 & 50.4 & 13.9 & 50.4 \\ 
        CenterNet  & 800  & o & 27.8 & 47.9 & 27.6 & 21.3 & 42.1 & 49.8 & 95.2 & 1855   \\
        HRDNet  & 1333  & o & 28.3 & 49.3 & 28.2 & - & - & - & - & 421 \\
        GFLV1 & 1333 & o & 28.4 & 50.0 & 27.8 & - & - & - & 525 & - \\
        CEASC  & 1333 & o & 28.7 & \textbf{50.7} & 28.4 & - & -& - & 43.8 & 150  \\ 
        \rowcolor{mycyan}
        {\bf RemDet-L} & 640  & o & 29.3 & 47.4 & 30.3 & 18.7 & 43.4 & 55.8 & 7.1 & 67.4 \\ 
        \rowcolor{mycyan}
        {\bf RemDet-X} & 640  & o & \textbf{29.9} & 48.3 & \textbf{31.0} & \textbf{19.5} & \textbf{44.1} & \textbf{58.6} & 8.9 & 114  \\ 
        \midrule
        ClusDet & 1000  & o+ca & 26.7 & 50.6 & 24.7 & 17.6 & 38.9 & 51.4 & 273 & -  \\
        DMNet & 1500  & o+ca  & 28.2 & 47.6 & 28.9 & 19.9 & 39.6 & 55.8 & 290 & - \\
        CDMNet & 1000  & ca  & 29.2 & 49.5 & 29.8 & 20.8 & 40.7 & 41.6 & - & -  \\
        GLASN & 600 & o+ca  & 30.7 & 55.4 & 30.0 & - & - & - & - & - \\
        AMRNet & 1500  & o+aug & 31.7 & - & - & 23.0 & 43.4 & \textbf{58.1} & - & - \\
        YOLC & 1024  & o+ca  & 31.8 & 55.0 & 31.7 & 24.7 & 42.3 & 45.0 & 441 & 151  \\
        CZDet & 1200  & o+dc & 33.2 & 58.3 & 33.2 & 26.0 & 42.6 & 43.4 & - & - \\
        UFPMP-Det & 1333  & o+ca & 36.6 & \textbf{62.4} & 36.7 & - & - & - & 152 & 205 \\
        \rowcolor{mycyan}
        {\bf RemDet-X} & 1024  & o+ca & \textbf{40.0} & 61.9 & \textbf{42.8} & \textbf{30.4} & \textbf{52.5} & 54.6 & 9.0 & 182 \\
		\bottomrule
	\end{tabular}
 }
  \caption {Comparison in terms of AP (\%), Latency, and FLOPs on VisDrone. o, ca, aug respectively stand for the original validation set, cluster-aware cropped images, and augmented images. "-" indicates that the result is not reported.}
\label{tab:table_remdet_compare_SOTA}
\end{table*}
\section{Experiment} 
\subsection{Experimental Setup}
\paragraph{Datasets}
To evaluate our method, we conduct UAV detection experiment on the VisDrone \cite{visdrone2019} and UAVDT \cite{uavdt}, and also included the MSCOCO \cite{mscoco} dataset as an additional benchmark. VisDrone comprises 8,599 aerial images across 10 categories, with 6,471 images for training and 548 images for validation, all at a resolution of 2,000$\times$1,500 pixels. Since the evaluation server is currently closed, we followed related works and used the validation set for performance evaluation. MSCOCO contains over 330,000 images with multiple annotations across 80 categories. UAVDT includes 23,258 training images and 15,069 testing images, with a resolution of 1,024$\times$540 pixels across 3 classes. 

\paragraph{Evaluation Measures}
The metric we use to evaluate and compare the performance of various methods is the COCO-style Average Precision (AP). Additionally, we report the average precision for small, medium, and large objects to assess our method’s performance in detecting small objects. Efficiency is represented using GFLOPs and latency.

\paragraph{Implementation Details}
Using PyTorch and MMDetection, we trained one-stage models from scratch on the VisDrone and UAVDT datasets for 300 epochs, with a learning rate of 1e-2, and applied data augmentation techniques such as mixup and Mosaic. For two-stage models, we utilized pre-trained backbone networks. On MSCOCO, we kept the same parameters, except for a momentum of 0.937, a weight decay of 5e-4, and a learning rate decay of 1e-2 every 10 epochs. The input size for the YOLO series models was 640$\times$640, while for other models it was 1,333$\times$800. All experiments were conducted on 8 NVIDIA RTX 4090 GPUs, with inference performed on a single 4090 GPU.

\subsection{Comparison with SOTA on UAV Datasets}
\begin{figure*}[!th]
    \centering
    \includegraphics[width=\linewidth]{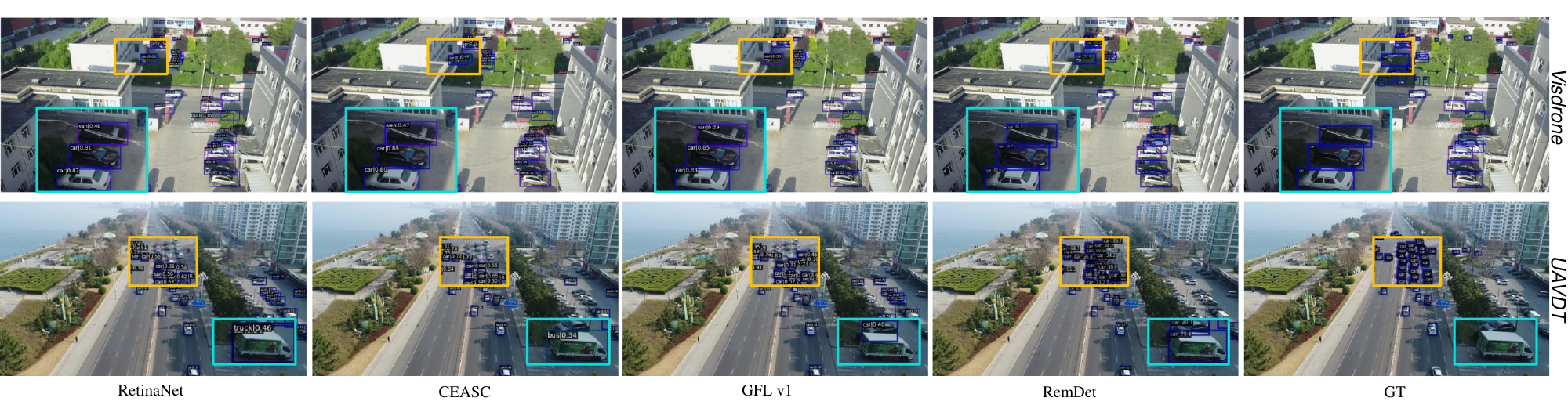}
    \caption{Visualization of the results of RetinaNet, CEASC, GFL and RemDet on the VisDrone and UAVDT datasets.}
    \label{fig:fig_vis_on_main_uavdatasets}
\end{figure*}
\begin{figure}[!th]
    \centering
    \includegraphics[width=\linewidth]{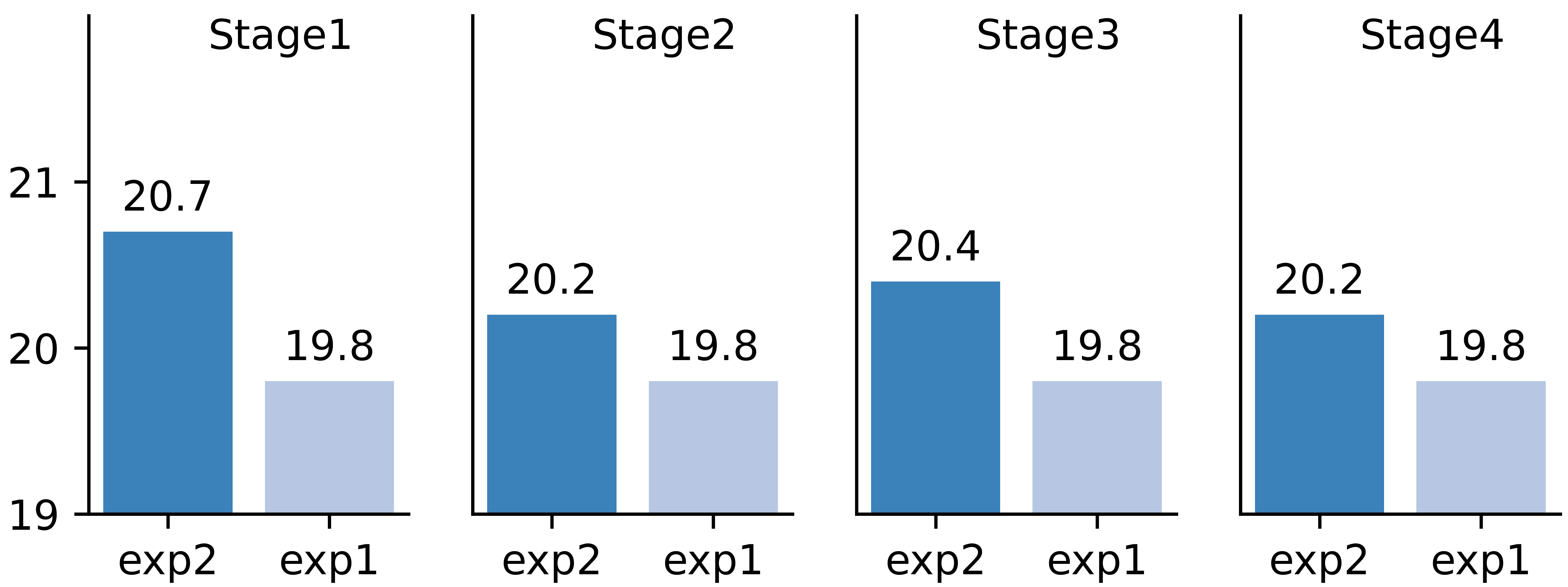}
    \caption{Ablation for each stage using different expansions.}
    \label{fig:fig_search_backbone}
\end{figure}

\paragraph{Comparison Results on VisDrone}
Our proposed model demonstrates significant improvements over existing models in terms of the key evaluation metric, mean Average Precision (mAP), on the VisDrone dataset. Additionally, to emphasize real-time performance, we compare our model with real-time general object detectors. Notably, the field of real-time lightweight UAV detection lacks comprehensive research. Our smallest model, RemDet-Tiny, achieves outstanding performance with an inference speed of 3.4 ms, surpassing the baseline \cite{yolov8} by 2.7\%.

When trained on the original validation set, as shown in Table \ref{tab:table_remdet_compare_SOTA}, our model outperforms the previous state-of-the-art lightweight model, CEASC \cite{ceasc}, by 0.8\%, while reducing computational complexity by 35\%. Compared to QueryDet \cite{querydet}, our model achieves a 9.6\% improvement. Furthermore, after incorporating the Cluster-Aware Crops method, our model aligns input image sizes with YOLC, achieving the best performance to date with an mAP of 40\%. This represents an 8.2\% improvement over YOLC and a 6.8\% improvement over CZDet. Remarkably, on a single 4090 GPU (without any model acceleration techniques), our model achieves a 9 ms inference speed, underscoring the effectiveness of our detector design.

However, our model shows a slight disadvantage in detecting large objects compared to other models. We attribute this to the intricate and heavily handcrafted designs used in competing methods, which can effectively detect large objects in UAV images. In contrast, our simpler design prioritizes high-speed inference capability and strong generalization.

\begin{table}[!tp]
	\centering	
    \resizebox{\linewidth}{!}{
	\begin{tabular}[c]{lcccccccc}
        \toprule
        Model & {AP$^{val}_{95}$} & {AP$^{val}_{50}$} & {AP$^{val}_{75}$} & {AP$^{val}_{s}$}  & {AP$^{val}_{m}$}  & {AP$^{val}_{l}$}\\
        \midrule
        R-FCN & 7.0 & 17.5 & 3.9 & 4.4 & 14.7 & 12.1\\
        FRCNN+FPN & 11.0 & 23.4 & 8.4 & 8.1 & 20.2 & 26.5\\
        CenterNet & 13.2 & 26.7 & 11.8 & 7.8 & 26.6 & 13.9\\
        ClusDet & 13.7 & 26.5 & 12.5 & 9.1 & 25.1 & 31.2 \\
        DMNet & 14.7 & 24.6 & 16.3 & 9.3 & 26.2 & 35.2 \\
        CDMNet & 16.8 & 29.1 & 18.5 &  11.9 & 29.0 & 15.7 \\
        GLSAN & 17.0 & 28.1 & 18.8 & - & - & -  \\
        CEASC & 17.1 & 30.9 & 17.8 & - & - & -  \\
        AMRNet & 18.2 & 30.4 & 19.8 & 10.3 & 31.3 & 33.5   \\
        YOLC & 19.3 & 30.9 & 20.1 & 10.9 & \textbf{32.2} & \textbf{35.5}   \\
        \rowcolor{mycyan}
        {\bf RemDet-L$^{\dagger}$} & \textbf{20.6} & \textbf{34.5} & \textbf{22.1} & \textbf{13.9} & 31.4 & 30.3  \\
        \bottomrule
    \end{tabular}
}
  \caption {Comparison in terms of AP (\%) on UAVDT. $^{\dagger}$ represents the use of cluster-aware cropped method.}
\label{tab:remdet_compare_uavdt}
 \end{table}
\paragraph{Comparison Results on UAVDT}
Based on our experimental results using the UAVDT dataset (as shown in Table \ref{tab:remdet_compare_uavdt}), we have demonstrated that the proposed method outperforms the current state-of-the-art model, YOLC, achieving the highest performance (20.6\%). Additionally, compared to other methods, our approach significantly improves the accuracy of small object detection by 3\%, although it performs worse for large objects. Furthermore, our methods can be combined with these handcrafted designs to obtain performance gains, even though it may introduce additional inference overhead.

\subsection{Ablation Study}
\paragraph{Ablation of Overall Design}
In our research, the block ratios for each stage are (3:6:6:3). This ratio was determined through neural architecture search (NAS) applied to the MSCOCO dataset. However, when dealing with UAV images, we aim to find the optimal block ratio by removing redundant modules for lightweight models.

To achieve this, we initially set all stage blocks to 3 and progressively increased them to 6. The experimental results are detailed in Appendix. Finally, We adopted a (3:3:6:3) block ratio, eliminating unnecessary blocks. In our CED module, the original channel expansion ratio was 1. Increasing it to 2 would enhance model performance, but at the cost of higher computation and latency. Thus, we aimed for a balance to maximize channel information expansion. Notably, channel expansion significantly improved stage 1 performance (Figure \ref{fig:fig_search_backbone}), leading us to remove it from other stages. This operation reduced model inference time to 3.7 ms, with only a 0.1\% performance drop. For detailed exploration of ours models, refer to Figure \ref{fig:fig_overall_design}.

\paragraph{More Ablation Experiments}
To validate our methods, we combine RemDet’s backbone with various detectors, as shown in Table \ref{table:evaluation on diff detectors}. Surprisingly, when combined with other detectors, our method consistently improves detection accuracy. Furthermore, we trained RemDet on the MSCOCO dataset (as shown in Table \ref{table:remdet_compare_mscoco}). In comparison to all minimal versions of generic detectors, our approach provides the most SOTA results, surpassing our baseline by 2.3\% on the COCO. We believe that our sufficiently simple structure contributes to powerful generalization, while also significantly enhancing small object detection accuracy on generic datasets.

\begin{table}
    \centering
\tabcolsep=0.35cm
    \resizebox{\linewidth}{!}{
        \begin{tabular}{l|l|lllll|rr}
            \toprule
           Base Detector & Backbone & {AP$^{val}_{95}$} & {AP$^{val}_{50}$} & {AP$^{val}_{s}$}  & {AP$^{val}_{m}$}  & {AP$^{val}_{l}$} & Param $\downarrow $ & FLOPs $\downarrow $ \\
           \midrule
            \multirow{2}{*}{YOLOv5-S} & CSPDarknet & 18.6 & 32.9 & 10.5 & 27.9 & 38.5 & 7.0M & 5.0G \\
            & \cellcolor{mycyan}RemDet & \cellcolor{mycyan}{\bf 20.7{ \scriptsize \textcolor{purple}{ $\uparrow$ 2.1}}} & \cellcolor{mycyan}{\bf 36.2{ \scriptsize \textcolor{purple}{ $\uparrow$ 3.3}}} &\cellcolor{mycyan}{\bf 12.1{ \scriptsize \textcolor{purple}{ $\uparrow$ \textbf{1.6}}}} &\cellcolor{mycyan}{\bf 31.1{ \scriptsize \textcolor{purple}{ $\uparrow$ \textbf{4.8}}}} &\cellcolor{mycyan}{\bf 40.7{ \scriptsize \textcolor{purple}{ $\uparrow$ \textbf{2.2}}}} & \cellcolor{mycyan}8.3M & \cellcolor{mycyan} 6.0G \\
            \midrule
            \multirow{2}{*}{RTMDet} & CSPNeXt & 21.8 & 35.2 & 11.5 & 35.2 & 49.5 & 8.9M & 9.3G \\
            & \cellcolor{mycyan}RemDet & \cellcolor{mycyan}{\bf 22.8{ \scriptsize \textcolor{purple}{ $\uparrow$ 1.0}}} & \cellcolor{mycyan}{\bf 35.8{ \scriptsize \textcolor{purple}{ $\uparrow$ 0.6}}} &\cellcolor{mycyan}{11.4{ \scriptsize \textcolor{gray}{ $\downarrow$ \textbf{0.1}}}} &\cellcolor{mycyan}{\bf 35.5{ \scriptsize \textcolor{purple}{ $\uparrow$ \textbf{0.3}}}} &\cellcolor{mycyan}{48.8{ \scriptsize \textcolor{gray}{ $\downarrow$ \textbf{0.7}}}} & \cellcolor{mycyan}9.9M & \cellcolor{mycyan} 10.1G \\
            \midrule
            \multirow{2}{*}{YOLOv8-S} & CSPDarknet & 23.1 & 38.7 & 13.9 & 34.8 & 42.3 & 11.1M & 14.3G \\
            & \cellcolor{mycyan}RemDet & \cellcolor{mycyan}{\bf 24.6{ \scriptsize \textcolor{purple}{ $\uparrow$ 1.5}}} & \cellcolor{mycyan}{\bf 41.3{ \scriptsize \textcolor{purple}{ $\uparrow$ 2.6}}} &\cellcolor{mycyan}{\bf 15.0{ \scriptsize \textcolor{purple}{ $\uparrow$ \textbf{1.1}}}} &\cellcolor{mycyan}{\bf 36.5{ \scriptsize \textcolor{purple}{ $\uparrow$ \textbf{1.7}}}} &\cellcolor{mycyan}{\bf 46.7{ \scriptsize \textcolor{purple}{ $\uparrow$ \textbf{4.5}}}} & \cellcolor{mycyan}11.5M & \cellcolor{mycyan} 14.7G \\
             \midrule
            \multirow{2}{*}{FasterRCNN (1$\times$)} & ResNet50 & 16.3 & 29.5 & 9.1 & 25.6 & 27.3  & 41.4M & 208G \\
            & \cellcolor{mycyan}RemDet & \cellcolor{mycyan}{\bf 19.0{ \scriptsize \textcolor{purple}{ $\uparrow$ 2.7}}} & \cellcolor{mycyan}\textbf{34.2}{ \scriptsize \textcolor{purple}{ $\uparrow$ \textbf{4.7}}} &\cellcolor{mycyan}{\bf 11.6{ \scriptsize \textcolor{purple}{ $\uparrow$ \textbf{2.5}}}} &\cellcolor{mycyan}{\bf 29.1{ \scriptsize \textcolor{purple}{ $\uparrow$ \textbf{3.5}}}} &\cellcolor{mycyan}{\bf 30.6{ \scriptsize \textcolor{purple}{ $\uparrow$ \textbf{3.3}}}} & \cellcolor{mycyan}39.1M & \cellcolor{mycyan}187G \\
             \midrule
            \multirow{2}{*}{RetinaNet (100e)} & ResNet50 & 8.0 & 14.6 & 3.3 & 14.3 & 18.4 & 36.5M & 210G \\
            & \cellcolor{mycyan}RemDet & \cellcolor{mycyan}{\bf 15.9{ \scriptsize \textcolor{purple}{ $\uparrow$ 7.9}}} & \cellcolor{mycyan}{\bf 27.7{ \scriptsize \textcolor{purple}{ $\uparrow$ 13.1}}} &\cellcolor{mycyan}{\bf 7.1{ \scriptsize \textcolor{purple}{ $\uparrow$ \textbf{3.8}}}} &\cellcolor{mycyan}{\bf 26.8{ \scriptsize \textcolor{purple}{ $\uparrow$ \textbf{12.5}}}} &\cellcolor{mycyan}{\bf 35.0{ \scriptsize \textcolor{purple}{ $\uparrow$ \textbf{16.6}}}} & \cellcolor{mycyan}31.9M & \cellcolor{mycyan} 190G \\
              \midrule
            \multirow{2}{*}{DyHead (2$\times$)} & ResNet50 & 13.8 & 24.6 & 7.1 & 22.5 & 25.5 & 38.9M & 110G \\
            & \cellcolor{mycyan}RemDet & \cellcolor{mycyan}{\bf 17.2{ \scriptsize \textcolor{purple}{ $\uparrow$ 3.4}}} & \cellcolor{mycyan}{\bf 29.1{ \scriptsize \textcolor{purple}{ $\uparrow$ 4.5}}} &\cellcolor{mycyan}{\bf 9.6{ \scriptsize \textcolor{purple}{ $\uparrow$ \textbf{2.5}}}} &\cellcolor{mycyan}{\bf 27.5{ \scriptsize \textcolor{purple}{ $\uparrow$ \textbf{5.0}}}} &\cellcolor{mycyan}{\bf 31.7{ \scriptsize \textcolor{purple}{ $\uparrow$ \textbf{6.2}}}} & \cellcolor{mycyan}36.7M & \cellcolor{mycyan} 91G \\
        
           \bottomrule
        \end{tabular}
        }
        \caption{Comparing AP(\%) and GFLOPs/Param of various detectors on VisDrone with our approach.}       
    \label{table:evaluation on diff detectors}
\end{table}
\begin{table}[!tp]
	\centering	
 
    \resizebox{\linewidth}{!}{
	\begin{tabular}[c]{l|ccccc|rr}
        \toprule
        Model & {AP$^{val}_{95}$} & {AP$^{val}_{50}$} & {AP$^{val}_{s}$}  & {AP$^{val}_{m}$}  & {AP$^{val}_{l}$} & Param & FLOPs  \\
        \midrule
        YOLOv6-3.0-N & 36.2 & 51.6 & 16.8 & 40.2 & 52.6  & 4.3M & 5.5G \\
        YOLOv8-N  & 37.3 & 52.6 & 18.8 & 41.0 & 53.5 & 3.0M & 4.1G \\
        YOLOv7-Tiny  & 37.5 & 55.8 & 19.9 & 41.1 & 50.8  & 5.2M & 6.2G \\
        \rowcolor{mycyan}
        {\bf RemDet-Tiny}  & \textbf{39.5} & \textbf{55.8} & \textbf{21.0} & \textbf{43.9} & \textbf{54.0} & 3.2M & 4.6G \\
       \midrule
       YOLO-MS-XS & 43.1 & 60.1 & 24.0 & 47.8 & 59.1 & 4.5M & 8.7G \\
       YOLOv6-v3.0-S & 43.7 & 60.8 & 23.6 & 48.7 & 59.8 & 17.2M & 21.9G \\ 
       YOLOv8-S   & 44.9 & 61.8 & 26.0 & 49.9 & \textbf{61.0} & 11.1M & 14.3G \\
       \rowcolor{mycyan} 
        {\bf RemDet-S}  & \textbf{45.5} & \textbf{62.8} & \textbf{27.8} & \textbf{50.5} & 60.0 & 11.9M & 16.0G \\
        \bottomrule
    \end{tabular}
}
\caption {Comparison of AP(\%) and params/FLOPs with the real-time approaches on MSCOCO.}
\label{table:remdet_compare_mscoco}
 \end{table}

\section{Conclusion}
\label{sec:conclusion}
In this paper, we present RemDet, a novel UAV object detector with a focus on small and dense objects in UAV imagery. Our approach involves designing modules that enhance inter-layer information while mitigating information loss. To meet real-time requirements, we explore cost-efficient multiplication operations, adopt reparameterization, and remove unnecessary components. Our experiments show that RemDet achieves state-of-the-art results and high-speed inference. However, further enhancements are needed for large objects. We hope this work inspires future research in UAV detectors.

\section{Acknowledgments}
This work was supported by the National Natural Science Foundation of China (62376252); Key Project of Natural Science Foundation of Zhejiang Province (LZ22F030003); Zhejiang Province Leading Geese Plan(2024C02G1123882).

\bibliography{aaai25}

\begin{thebibliography}{36}
\providecommand{\natexlab}[1]{#1}

\bibitem[{Ashraf, Sultani, and Shah(2021)}]{dogfight}
Ashraf, M.~W.; Sultani, W.; and Shah, M. 2021.
\newblock Dogfight: Detecting drones from drones videos.
\newblock In \emph{Proceedings of the IEEE/CVF Conference on Computer Vision and Pattern Recognition}, 7067--7076.

\bibitem[{Carion et~al.(2020)Carion, Massa, Synnaeve, Usunier, Kirillov, and Zagoruyko}]{detr}
Carion, N.; Massa, F.; Synnaeve, G.; Usunier, N.; Kirillov, A.; and Zagoruyko, S. 2020.
\newblock End-to-end object detection with transformers.
\newblock In \emph{European conference on computer vision}, 213--229. Springer.

\bibitem[{Dauphin et~al.(2017)Dauphin, Fan, Auli, and Grangier}]{glu}
Dauphin, Y.~N.; Fan, A.; Auli, M.; and Grangier, D. 2017.
\newblock Language modeling with gated convolutional networks.
\newblock In \emph{International conference on machine learning}, 933--941. PMLR.

\bibitem[{Ding et~al.(2018)Ding, Xue, Long, Xia, and Lu}]{roitransformer}
Ding, J.; Xue, N.; Long, Y.; Xia, G.-S.; and Lu, Q. 2018.
\newblock Learning RoI transformer for detecting oriented objects in aerial images.
\newblock \emph{arXiv preprint arXiv:1812.00155}.

\bibitem[{Ding et~al.(2021)Ding, Zhang, Ma, Han, Ding, and Sun}]{ding2021repvgg}
Ding, X.; Zhang, X.; Ma, N.; Han, J.; Ding, G.; and Sun, J. 2021.
\newblock Repvgg: Making vgg-style convnets great again.
\newblock In \emph{Proceedings of the IEEE/CVF conference on computer vision and pattern recognition}, 13733--13742.

\bibitem[{Dosovitskiy et~al.(2020)Dosovitskiy, Beyer, Kolesnikov, Weissenborn, Zhai, Unterthiner, Dehghani, Minderer, Heigold, Gelly et~al.}]{vit}
Dosovitskiy, A.; Beyer, L.; Kolesnikov, A.; Weissenborn, D.; Zhai, X.; Unterthiner, T.; Dehghani, M.; Minderer, M.; Heigold, G.; Gelly, S.; et~al. 2020.
\newblock An image is worth 16x16 words: Transformers for image recognition at scale.
\newblock \emph{arXiv preprint arXiv:2010.11929}.

\bibitem[{Du et~al.(2023)Du, Huang, Chen, and Huang}]{ceasc}
Du, B.; Huang, Y.; Chen, J.; and Huang, D. 2023.
\newblock Adaptive sparse convolutional networks with global context enhancement for faster object detection on drone images.
\newblock In \emph{Proceedings of the IEEE/CVF conference on computer vision and pattern recognition}, 13435--13444.

\bibitem[{Du et~al.(2018)Du, Qi, Yu, Yang, Duan, Li, Zhang, Huang, and Tian}]{uavdt}
Du, D.; Qi, Y.; Yu, H.; Yang, Y.; Duan, K.; Li, G.; Zhang, W.; Huang, Q.; and Tian, Q. 2018.
\newblock The unmanned aerial vehicle benchmark: Object detection and tracking.
\newblock In \emph{Proceedings of the European conference on computer vision (ECCV)}, 370--386.

\bibitem[{Duan et~al.(2019)Duan, Bai, Xie, Qi, Huang, and Tian}]{centernet}
Duan, K.; Bai, S.; Xie, L.; Qi, H.; Huang, Q.; and Tian, Q. 2019.
\newblock Centernet: Keypoint triplets for object detection.
\newblock In \emph{Proceedings of the IEEE/CVF international conference on computer vision}, 6569--6578.

\bibitem[{Girshick(2015)}]{fasterrcnn}
Girshick, R. 2015.
\newblock Fast r-cnn.
\newblock In \emph{Proceedings of the IEEE international conference on computer vision}, 1440--1448.

\bibitem[{Han et~al.(2021)Han, Yun, Heo, and Yoo}]{rethinkingchannel}
Han, D.; Yun, S.; Heo, B.; and Yoo, Y. 2021.
\newblock Rethinking channel dimensions for efficient model design.
\newblock In \emph{Proceedings of the IEEE/CVF conference on Computer Vision and Pattern Recognition}, 732--741.

\bibitem[{{Hollard} et~al.(2024){Hollard}, {Mohimont}, {Gaveau}, and {Steffenel}}]{leyolo}
{Hollard}, L.; {Mohimont}, L.; {Gaveau}, N.; and {Steffenel}, L.-A. 2024.
\newblock {LeYOLO, New Scalable and Efficient CNN Architecture for Object Detection}.
\newblock \emph{arXiv e-prints}, arXiv:2406.14239.

\bibitem[{Howard et~al.(2019)Howard, Sandler, Chu, Chen, Chen, Tan, Wang, Zhu, Pang, Vasudevan et~al.}]{mobilenetv3}
Howard, A.; Sandler, M.; Chu, G.; Chen, L.-C.; Chen, B.; Tan, M.; Wang, W.; Zhu, Y.; Pang, R.; Vasudevan, V.; et~al. 2019.
\newblock Searching for mobilenetv3.
\newblock In \emph{Proceedings of the IEEE/CVF international conference on computer vision}, 1314--1324.

\bibitem[{Huang, Chen, and Huang(2022)}]{ufpmpdet}
Huang, Y.; Chen, J.; and Huang, D. 2022.
\newblock UFPMP-Det: Toward accurate and efficient object detection on drone imagery.
\newblock In \emph{Proceedings of the AAAI conference on artificial intelligence}, volume~36, 1026--1033.

\bibitem[{Jocher(2020)}]{yolov5}
Jocher, G. 2020.
\newblock {YOLOv5 by Ultralytics}.

\bibitem[{Jocher, Chaurasia, and Qiu(2023)}]{yolov8}
Jocher, G.; Chaurasia, A.; and Qiu, J. 2023.
\newblock {Ultralytics YOLO}.

\bibitem[{Li et~al.(2022)Li, Li, Jiang, Weng, Geng, Li, Ke, Li, Cheng, Nie et~al.}]{yolov6}
Li, C.; Li, L.; Jiang, H.; Weng, K.; Geng, Y.; Li, L.; Ke, Z.; Li, Q.; Cheng, M.; Nie, W.; et~al. 2022.
\newblock YOLOv6: A single-stage object detection framework for industrial applications.
\newblock \emph{arXiv preprint arXiv:2209.02976}.

\bibitem[{Lin et~al.(2014)Lin, Maire, Belongie, Hays, Perona, Ramanan, Doll{\'a}r, and Zitnick}]{mscoco}
Lin, T.-Y.; Maire, M.; Belongie, S.; Hays, J.; Perona, P.; Ramanan, D.; Doll{\'a}r, P.; and Zitnick, C.~L. 2014.
\newblock Microsoft coco: Common objects in context.
\newblock In \emph{Computer Vision--ECCV 2014: 13th European Conference, Zurich, Switzerland, September 6-12, 2014, Proceedings, Part V 13}, 740--755. Springer.

\bibitem[{Liu et~al.(2024)Liu, Gao, Huang, Hu, Liu, and Wang}]{yolc}
Liu, C.; Gao, G.; Huang, Z.; Hu, Z.; Liu, Q.; and Wang, Y. 2024.
\newblock YOLC: You Only Look Clusters for Tiny Object Detection in Aerial Images.
\newblock \emph{IEEE Transactions on Intelligent Transportation Systems}.

\bibitem[{Liu et~al.(2023)Liu, Peng, Zheng, Yang, Hu, and Yuan}]{efficientvit}
Liu, X.; Peng, H.; Zheng, N.; Yang, Y.; Hu, H.; and Yuan, Y. 2023.
\newblock Efficientvit: Memory efficient vision transformer with cascaded group attention.
\newblock In \emph{Proceedings of the IEEE/CVF Conference on Computer Vision and Pattern Recognition}, 14420--14430.

\bibitem[{Liu et~al.(2022)Liu, Mao, Wu, Feichtenhofer, Darrell, and Xie}]{convnext}
Liu, Z.; Mao, H.; Wu, C.-Y.; Feichtenhofer, C.; Darrell, T.; and Xie, S. 2022.
\newblock A convnet for the 2020s.
\newblock In \emph{Proceedings of the IEEE/CVF conference on computer vision and pattern recognition}, 11976--11986.

\bibitem[{Meethal, Granger, and Pedersoli(2023)}]{czdet}
Meethal, A.; Granger, E.; and Pedersoli, M. 2023.
\newblock Cascaded Zoom-in Detector for High Resolution Aerial Images.
\newblock In \emph{Proceedings of the IEEE/CVF Conference on Computer Vision and Pattern Recognition}, 2045--2054.

\bibitem[{Redmon et~al.(2016)Redmon, Divvala, Girshick, and Farhadi}]{yolov1}
Redmon, J.; Divvala, S.; Girshick, R.; and Farhadi, A. 2016.
\newblock You only look once: Unified, real-time object detection.
\newblock In \emph{Proceedings of the IEEE conference on computer vision and pattern recognition}, 779--788.

\bibitem[{Redmon and Farhadi(2017)}]{yolov2}
Redmon, J.; and Farhadi, A. 2017.
\newblock YOLO9000: better, faster, stronger.
\newblock In \emph{Proceedings of the IEEE conference on computer vision and pattern recognition}, 7263--7271.

\bibitem[{Sandler et~al.(2018)Sandler, Howard, Zhu, Zhmoginov, and Chen}]{mobilenetv2}
Sandler, M.; Howard, A.; Zhu, M.; Zhmoginov, A.; and Chen, L.-C. 2018.
\newblock Mobilenetv2: Inverted residuals and linear bottlenecks.
\newblock In \emph{Proceedings of the IEEE conference on computer vision and pattern recognition}, 4510--4520.

\bibitem[{Shwartz-Ziv and Tishby(2017)}]{shwartz2017opening}
Shwartz-Ziv, R.; and Tishby, N. 2017.
\newblock Opening the black box of deep neural networks via information.
\newblock \emph{arXiv preprint arXiv:1703.00810}.

\bibitem[{Tishby and Zaslavsky(2015)}]{tishby2015deep}
Tishby, N.; and Zaslavsky, N. 2015.
\newblock Deep learning and the information bottleneck principle.
\newblock In \emph{2015 ieee information theory workshop (itw)}, 1--5. IEEE.

\bibitem[{Wang et~al.(2024)Wang, Chen, Lin, Han, and Ding}]{repvitwang2024}
Wang, A.; Chen, H.; Lin, Z.; Han, J.; and Ding, G. 2024.
\newblock Repvit: Revisiting mobile cnn from vit perspective.
\newblock In \emph{Proceedings of the IEEE/CVF Conference on Computer Vision and Pattern Recognition}, 15909--15920.

\bibitem[{Wang, Bochkovskiy, and Liao(2023)}]{yolov7wang2023}
Wang, C.-Y.; Bochkovskiy, A.; and Liao, H.-Y.~M. 2023.
\newblock YOLOv7: Trainable bag-of-freebies sets new state-of-the-art for real-time object detectors.
\newblock In \emph{Proceedings of the IEEE/CVF Conference on Computer Vision and Pattern Recognition}, 7464--7475.

\bibitem[{Wang, Yeh, and Liao(2024)}]{yolov9}
Wang, C.-Y.; Yeh, I.-H.; and Liao, H.-Y.~M. 2024.
\newblock YOLOv9: Learning What You Want to Learn Using Programmable Gradient Information.
\newblock \emph{arXiv preprint arXiv:2402.13616}.

\bibitem[{Wang et~al.(2020)Wang, Wu, Zhu, Li, Zuo, and Hu}]{wang2020eca}
Wang, Q.; Wu, B.; Zhu, P.; Li, P.; Zuo, W.; and Hu, Q. 2020.
\newblock ECA-Net: Efficient channel attention for deep convolutional neural networks.
\newblock In \emph{Proceedings of the IEEE/CVF conference on computer vision and pattern recognition}, 11534--11542.

\bibitem[{Wei et~al.(2020)Wei, Duan, Song, Tian, and Wang}]{amrnet}
Wei, Z.; Duan, C.; Song, X.; Tian, Y.; and Wang, H. 2020.
\newblock Amrnet: Chips augmentation in aerial images object detection.
\newblock \emph{arXiv preprint arXiv:2009.07168}.

\bibitem[{Yang, Huang, and Wang(2022)}]{querydet}
Yang, C.; Huang, Z.; and Wang, N. 2022.
\newblock QueryDet: Cascaded sparse query for accelerating high-resolution small object detection.
\newblock In \emph{Proceedings of the IEEE/CVF Conference on computer vision and pattern recognition}, 13668--13677.

\bibitem[{Yang et~al.(2019)Yang, Fan, Chu, Blasch, and Ling}]{yang2019clustered}
Yang, F.; Fan, H.; Chu, P.; Blasch, E.; and Ling, H. 2019.
\newblock Clustered object detection in aerial images.
\newblock In \emph{Proceedings of the IEEE/CVF international conference on computer vision}, 8311--8320.

\bibitem[{Ye et~al.(2023)Ye, Zhang, Chen, Fan, and Wang}]{pagcp}
Ye, H.; Zhang, B.; Chen, T.; Fan, J.; and Wang, B. 2023.
\newblock Performance-aware approximation of global channel pruning for multitask cnns.
\newblock \emph{IEEE Transactions on Pattern Analysis and Machine Intelligence}, 45(8): 10267--10284.

\bibitem[{Zhu et~al.(2021)Zhu, Wen, Du, Bian, Fan, Hu, and Ling}]{visdrone2019}
Zhu, P.; Wen, L.; Du, D.; Bian, X.; Fan, H.; Hu, Q.; and Ling, H. 2021.
\newblock Detection and tracking meet drones challenge.
\newblock \emph{IEEE Transactions on Pattern Analysis and Machine Intelligence}, 44(11): 7380--7399.

\end{thebibliography}

\clearpage

\appendix
\section{Supplementary Material}
\label{sec:appendix_section}

\subsection{A. Training Recipes}
\begin{table}[!h]
    \centering
    \resizebox{\linewidth}{!}{
        \begin{tabular}{c|c}
            \specialrule{.08em}{0pt} {.10ex}
            config & object detection \\
            \hline
            optimizer & SGD\\
            base learning rate & 0.01 \\
            weight decay & 0.0005 \\
            optimizer momentum & 0.937 \\
            batch size & 128 \\
            learning rate schedule & Flat-Cosine \\
            training & 300 \\
            input size & 640 $\times$ 640 \\
            augmentation & Mosaic and MixUp \\
            EMA decay & 0.9998 \\
            \specialrule{.08em}{0ex}{0pt}
        \end{tabular}
    }
    \caption{Training settings for object detection.}
    \label{table:train_recipes}
\end{table}
The training parameters we used are from MMYOLO. To align with other works, we integrated the MMYOLO project into MMDetection. This paper details the parameter settings used during training. The optimizer used is SGD with a learning rate set to 0.01. Due to training with 8 GPUs, the training accuracy of the smallest model is relatively low. Therefore, we multiplied the learning rate by the number of GPUs to align the training accuracy, while keeping the learning rate at 0.01 for other models. Additionally, the batch size is set to 128. For other parameters, we used Mosaic and MixUp from the YOLO series as data augmentation techniques.

\subsection{B. The design of ConvFFN and Multiplication}
In this section, we present a detailed description of the design for ConvFFN and Multiplication. ConvFFN comprises two 1$\times$1 convolutional layers, which structurally resemble a feed-forward neural network. In contrast, Multiplication builds upon ConvFFN by splitting the input $x$. One part of the input undergoes forgetting through a non-linear activation, while the main portion is processed by an activation function. Finally, these two components are multiplied and output through a 1$\times$1 convolutional layer. For more detailed PyTorch code examples, please refer to list \ref{alg:algorithm_convffn} and list \ref{alg:algorithm_multiplication}.

\begin{figure}[!h]
    \centering
    \includegraphics[width=\linewidth]{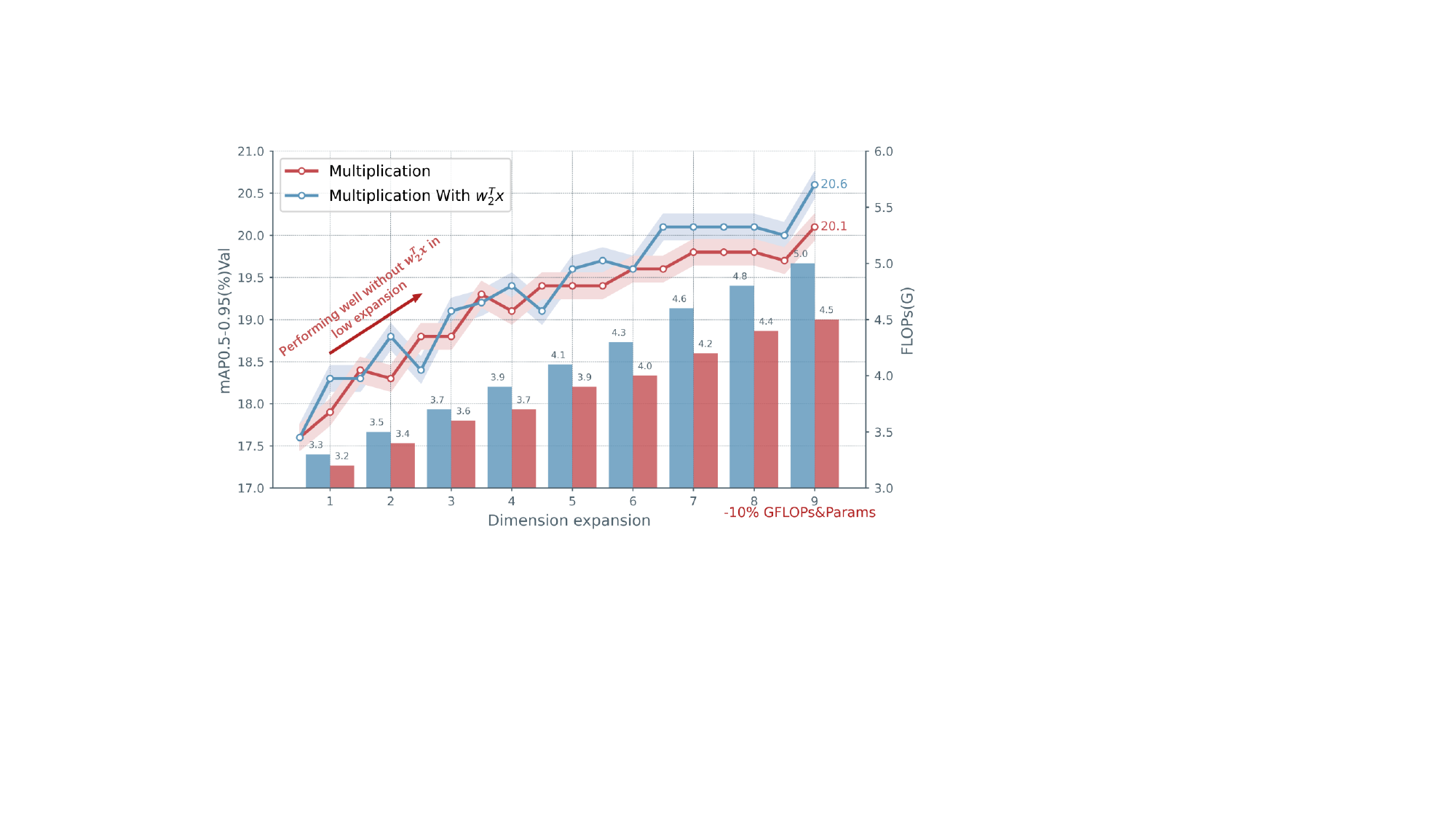}
    \caption{Compare Multiplication and with $w^T_2x$.}
    \label{fig:withx2_multiplication}
\end{figure}

To further supplement what was described in Section 3.2, namely that Multiplication discards the $w_2^Tx$ part, we compress it after the multiplication ends together with $w_1^Tx$ into the output dimension. The results, as shown in Figure \ref{fig:withx2_multiplication}, first retain $w_2^Tx$ , which has a similar form to MLP but significantly higher accuracy than MLP. In the case of an expansion factor of 9, the accuracy improves by 0.4\%. Notably, Multiplication introduces negligible computational overhead. Furthermore, Multiplication continues to perform strongly even in low channel expansion. We believe this is because after multiplication, the high-dimensional polynomial reduces the number of terms through addition, implicitly encoding information from $w_2^Tx$, while the performance loss is 2\% (when the expansion is 9). This result further underscores the advantage of multiplication over MLP.

\begin{figure}[!h]
    \centering
    \includegraphics[width=\linewidth]{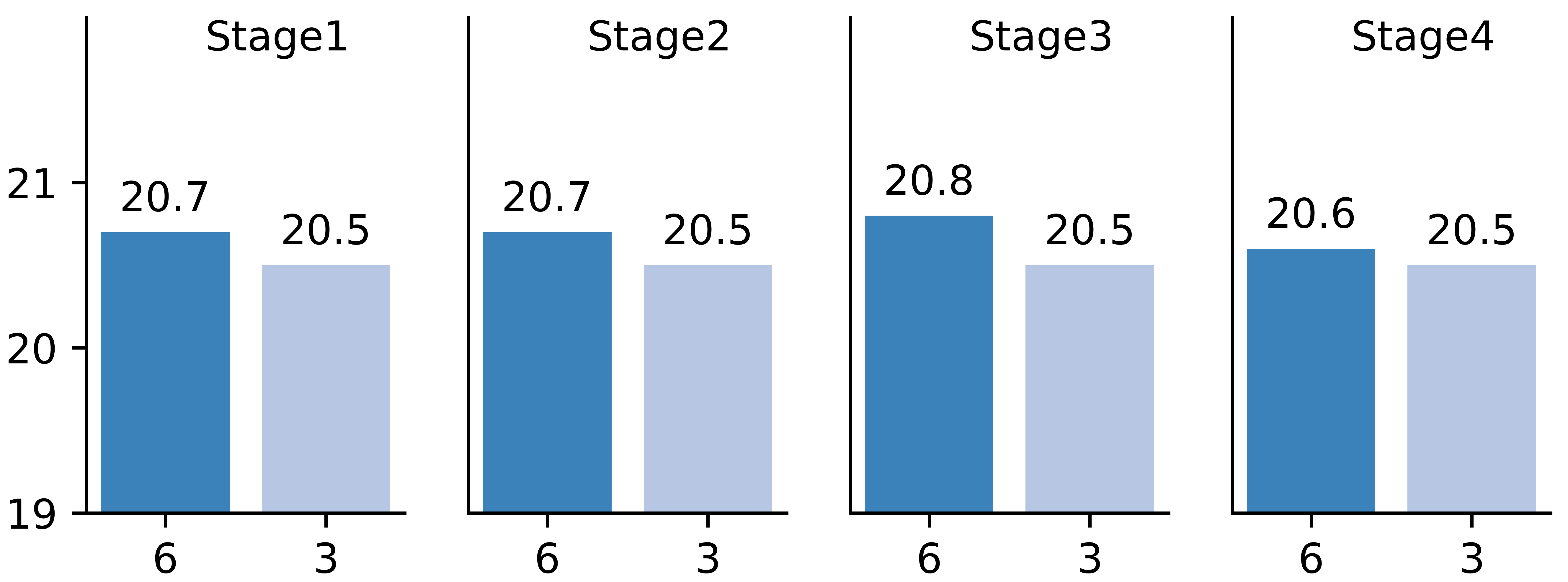}
    \caption{Ablation of Backbone's stage ratios.}
    \label{fig:the_ablation_backbone_stage_ratios}
\end{figure}
\subsection{C. Ablation Experiments of RemDet}
When conducting ablation studies on the VisDrone dataset, we analyzed the impact of each component on detection accuracy and computational complexity, as shown in Table \ref{table:ablation RemNet}. We separately trained GatedFFN, CED, and ChannelC2f. Compared to the baseline, the mAP improved by 1.2\%, 0.7\%, and 0.8\%, respectively. When combining GatedFFN and CED, the improvement was even greater, reaching 2.4\%. Notably, in the ablation experiments, CED had a fixed channel expansion setting of 1 across all stages. Furthermore, by incorporating ChannelC2f into the Neck layer, we achieved a further accuracy boost, reaching 21.8\%, while the computational cost increased by only 0.6 GFLOPs compared to the baseline.

\subsection{D. Compare Real-time Object Detectors on VisDrone}
We compared RemDet with real-time general object detectors on the VisDrone dataset, conducting experiments across five different scales. Due to the lack of support for some of the latest detectors in mmdetection, we faced difficulties in conducting comprehensive experiments on VisDrone.

As shown in Table \ref{tab:appendix_remdet_compare_general_rt_detectors}, our model outperformed existing real-time detectors at all scales. Notably, RemDet-M achieved the greatest performance improvement (an increase of 3.8\%) compared to other real-time detectors. Another noteworthy observation is that RemDet exhibits a higher computational load compared to the baseline at smaller scales. However, as the model scale increases, our approach significantly reduces the computational burden. This phenomenon can be attributed to our channel expansion. As the model scales up, it naturally increases channel dimensions and involves more dense computation. Our method effectively mitigates this dense computation, resulting in a wider overall model. Importantly, the results demonstrate that at larger scales, our approach not only maintains a precision advantage but also achieves greater efficiency due to its reduced computational overhead.

\subsection{E. The Design of CED}
\begin{figure}[!tp]
    \centering
    \includegraphics[width=\linewidth]{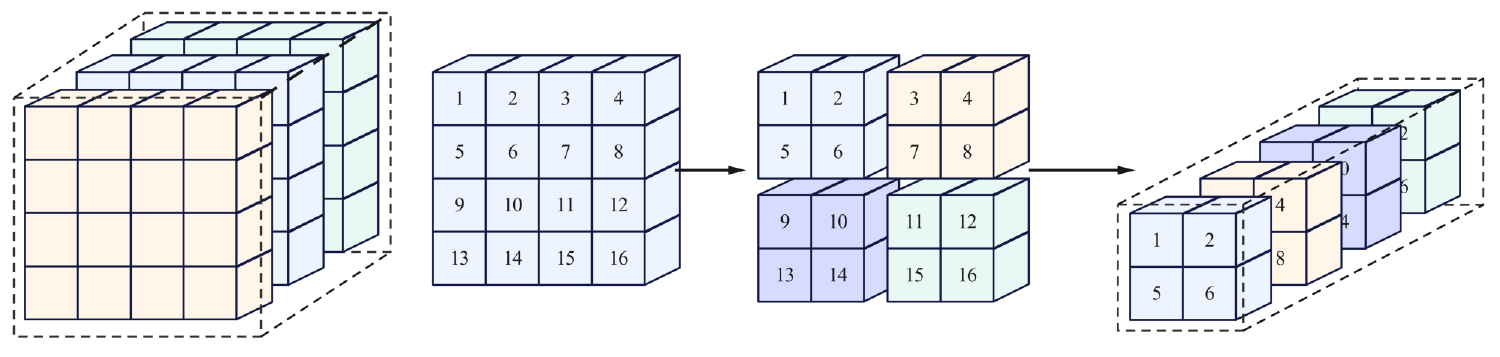}
    \caption{The detailed designs of CED module.}
    \label{fig:the_design_of_ced}
\end{figure}

As shown in Figure \ref{fig:the_design_of_ced}, the detailed design of CED involves combining different channels. Each color represents a distinct channel that is concatenated together. Taking an example channel, we split it into four subgraphs and then concatenate them along the channel dimension. This operation, described as the ‘patch merge’ process in the paper, results in the feature map being reduced to half its original size while increasing the channel dimension by a factor of four. For pointwise convolutions, this approach effectively expands the content that can be modeled by a factor of four, including some information from the surrounding pixels in the original feature map, enhancing information interaction.

\subsection{F. The Stage Ratios of Backbone}
When investigating the contribution of the number of blocks in each stage of the detector to overall performance, we conducted a series of ablation experiments. Starting from the baseline model, we used only GatedFFN and employed default convolutional downsampling. Next, we set the block proportions for each stage to 3 and gradually increased the size of each stage, adjusting the proportions from 3 to 6. The experimental results are shown in Figure \ref{fig:the_ablation_backbone_stage_ratios}. These results clearly indicate that the third stage has the greatest impact on the overall model performance. Notably, when maintaining the previous proportions (3:6:6:3), the accuracy was only 20.3\%. Consequently, we removed the redundant second-stage model and adjusted the proportions to (3:3:6:3).

\subsection{G. More Visualization of RemDet}
\begin{figure*}[!tp]
    \centering
    \includegraphics[width=\linewidth]{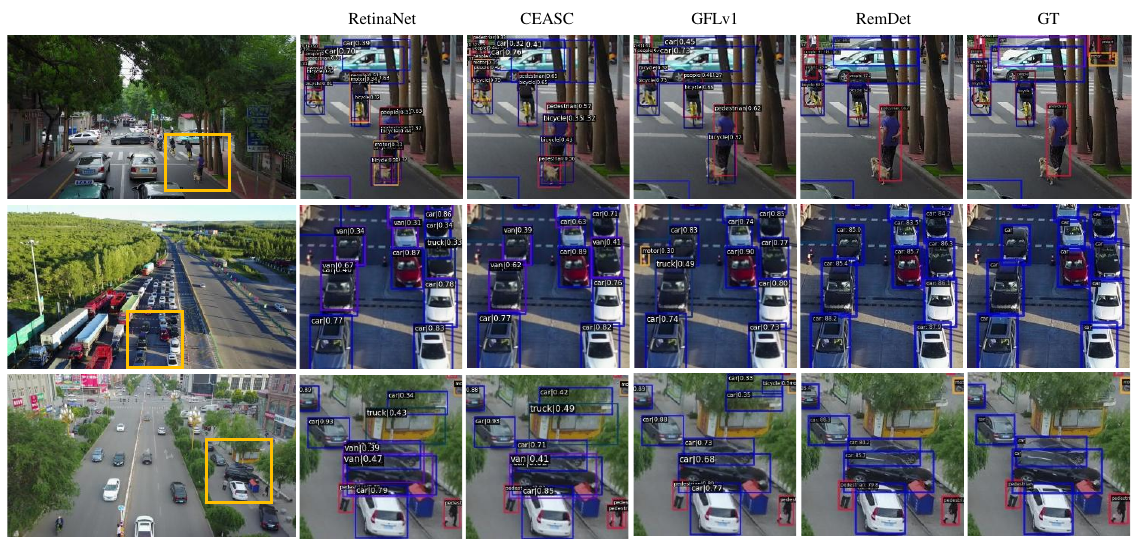}
    \caption{More visualized results of RetinaNet, CEASC, GFL and RemDet on the VisDrone.}
    \label{fig:more_visualize}
\end{figure*}

We conducted an extensive evaluation of object detection methods, including RetinaNet, CEASC, GFLv1, and RemDet, using the VisDrone dataset. Figure \ref{fig:more_visualize} illustrates our findings. In the topmost image, our approach stands out by accurately identifying non-detection objects, while other methods mistakenly classify pets as pedestrians or bicycles. This highlights the superior robustness of our method. Moving to the middle image, even in densely detected vehicle scenes with varying colors and vehicle types, our approach consistently identifies vehicle categories. Lastly, in the bottom image, our method not only avoids misidentifying street-side objects but also correctly recognizes different vehicle types.

\subsection{H. Compare with More Detectors on MSCOCO}
We compared our detector with several detectors on the MSCOCO dataset, as shown in Table \ref{table:remdet_m_mscoco}. YOLOv10 is built using the ultralytics library, but its performance is not extensively reported in the paper. On the other hand, GOLD-YOLO calculates parameter count and computational cost using the thop library, but there is no fair comparison with our fvcore-based method. However, in the case of the tiny version, RemDet-Tiny outperforms YOLOv10-N by 1.0\%. It’s important to note that our model demonstrates superiority only in the tiny version, revealing limitations in accuracy for larger versions. In other versions, we supplement the performance of the M model, which was not reported in the main text. Compared to some state-of-the-art detectors like GOLD-YOLO, RemDet shows a comparable performance gap, and despite not reporting parameter count and computational cost, RemDet exhibits significantly faster inference speed than GOLD-YOLO.

\begin{table}[!tp]
    \centering
    \resizebox{\linewidth}{!}{
        \begin{tabular}{cccllrr}
            \toprule
            GatedFFN & CED & ChannelC2f & {AP$^{val}_{95}$} & AP$^{val}_{50}$ & Param$\downarrow$ & FLOPs$\downarrow$ \\
            \midrule
              &  &  &   19.1 & 33.0 & 3.0M & 4.1G\\
            \checkmark & & & 20.3{ \scriptsize \textcolor{purple}{ $\uparrow$ 1.2}}& 34.9{ \scriptsize \textcolor{purple}{ $\uparrow$ 1.9}} & 3.9M & 5.1G\\
            & \checkmark & & 19.8{ \scriptsize \textcolor{purple}{ $\uparrow$ 0.7}} & 34.3{ \scriptsize \textcolor{purple}{ $\uparrow$ 1.3}} & 3.0M & 4.2G\\
            & & \checkmark & 19.9{ \scriptsize \textcolor{purple}{ $\uparrow$ 0.8}} & 33.9{ \scriptsize \textcolor{purple}{ $\uparrow$ 0.9}} & 3.4M & 4.5G\\
            \checkmark & \checkmark &  & 21.5{ \scriptsize \textcolor{purple}{ $\uparrow$ 2.4}} & 36.7{ \scriptsize \textcolor{purple}{ $\uparrow$ 3.7}} & 3.1M & 4.3G \\
            \checkmark & \checkmark & \checkmark & 21.7{ \scriptsize \textcolor{purple}{ $\uparrow$ 2.6}} & 36.9{ \scriptsize \textcolor{purple}{ $\uparrow$ 3.9}} & 3.5M & 4.6G \\
            \bottomrule
        \end{tabular}
    }
    \caption{Ablation on RemDet-Tiny with YOLOv8-N as the base detector on VisDrone.}
    \label{table:ablation RemNet}
\end{table}
\begin{table}[!tp]
	\centering	
    \resizebox{\linewidth}{!}{
	\begin{tabular}[c]{l|ccccc|rr}
        \toprule
        Model & {AP$^{val}_{95}$} & {AP$^{val}_{50}$} & {AP$^{val}_{s}$}  & {AP$^{val}_{m}$}  & {AP$^{val}_{l}$} & Param & FLOPs  \\
        \midrule
        YOLOv10-N  & 38.5 & - & - & - & - & - & - \\
        \rowcolor{mycyan}
        {\bf RemDet-Tiny}  & \textbf{39.5} & \textbf{55.8} & \textbf{21.0} & \textbf{43.9} & \textbf{54.0} & 3.2M & 4.6G \\
       \midrule
       RTMDet-S & 44.6 & 61.7 & 24.2 & 49.2 & 61.8 & 8.9M & 14.8G \\ 
       GOLD-YOLO-S & 45.4 & 62.5 & 25.3 & 50.2 & \textbf{62.6} & - & - \\
       \rowcolor{mycyan} 
        {\bf RemDet-S}  & \textbf{45.5} & \textbf{62.8} & \textbf{27.8} & \textbf{50.5} & 60.0 & 11.9M & 16.0G \\
         \midrule
         YOLOv6-3.0-M & 48.4 & 65.7 & 30.0 & 54.1 & 64.5 & 34.3M & 40.7G \\ 
         RTMDet-M & 49.3 & 66.9 & 30.5 & 53.6 & 66.1 & 24.7M & 39.2G \\ 
         DAMO YOLO-M & 49.2 & 65.5 & 29.7 & 53.1 & 66.1 & - & - \\ 
         GOLD-YOLO-M & 49.8 &\bf 67.0 & 32.3 &\bf 55.3 &\bf 66.3 & - & - \\ 
          \rowcolor{mycyan} 
        {\bf RemDet-M}  & \textbf{49.8} & 66.9 & \textbf{32.8} & 54.7 & 65.0 & 23.3M & 34.4G \\
        \bottomrule
    \end{tabular}
}
\caption {Comparison with the real-time approaches on MSCOCO.}
\label{table:remdet_m_mscoco}
 \end{table}

\begin{table*}[!tp]
	\centering	
\scalebox{0.9}{
	\begin{tabular}[c]{l|ccc|ccc|cc}
		\toprule
		Model & {AP$^{val}_{95}$} & {AP$^{val}_{50}$} & {AP$^{val}_{75}$}& {AP$^{val}_{s}$}  & {AP$^{val}_{m}$}  & {AP$^{val}_{l}$} & {Param(M)$\downarrow$} & {FLOPs(G)$\downarrow$}  \\
		\midrule
		YOLOv5-N & 14.4 & 26.7 & 13.7 & 6.8 & 22.4 & 32.6 & 1.8 & 1.3 \\
        YOLOv6-v3.0-N & 19.0 & 32.8 & 18.7 & 9.9 & 29.0 & 41.3 & 5.0 & 3.9 \\
        YOLOv7-Tiny & 19.4 & 35.1 & 18.5 & 10.5 & 29.1 & 41.0 & 6.0 & 4.2 \\
        YOLOv8-N & 19.1 & 33.0 & 18.9 & 10.6 & 28.9 & 38.3 & 3.0 & 4.1 \\
        RTMDet-Tiny & 20.3 & 33.5 & 21.2 & 10.2 & 32.9 & \textbf{47.1} & 4.9 & 5.1 \\
        \rowcolor{mycyan}
        {\bf RemDet-Tiny} & {\bf 21.8} & {\bf 37.1} & \textbf{21.9} & \textbf{12.7} & \textbf{33.0} & 44.5 & 3.2 & 4.6 \\
        \midrule
        YOLOX-S & 14.9 & 29.0 & 13.6 & 8.1 & 22.8 & 28.1 & 8.9 & 13.3 \\
        YOLOv5-S & 18.6 & 32.8 & 18.3 & 10.5 & 28.1 & 36.7 & 7.0 & 5.0 \\
        RTMDet-S & 21.9 & 35.5 & 22.9 & 11.7 & 35.3 & \textbf{49.4} & 8.9 & 9.3 \\
        YOLOv8-S & 23.2 & 39.0 & 23.3 & 13.8 & 34.8 & 44.9 & 11.1 & 14.3 \\
        YOLOv6-v3.0-S & 23.4 & 39.4 & 23.6 & 13.6 & 35.2 & 44.8 & 20.1 & 15.5 \\
        \rowcolor{mycyan}
        {\bf RemDet-S} & \textbf{24.7} & \textbf{41.5} & \textbf{25.0} & \textbf{15.4} & \textbf{36.7} & 47.0 & 11.9 & 16.0 \\
        \midrule
        YOLOX-M & 18.6 & 34.4 & 17.6 & 11.0 & 28.0 & 34.4 & 25.3 & 36.8 \\
        YOLOv5-M & 22.3 & 37.9 & 22.4 & 12.5 & 33.8 & 47.1 & 20.9 & 15.2 \\
        RTMDet-M & 23.0 & 36.3 & 24.5 & 11.9 & 37.4 & 49.5 & 24.7 & 24.6 \\
        YOLOv6-v3.0-M & 24.1 & 40.3 & 24.5 & 14.2 & 36.3 & 45.6 & 37.7 & 29.2 \\
        YOLOv8-M & 24.4 & 40.5 & 24.8 & 14.5 & 36.7 & 46.0 & 25.9 & 39.5 \\
        \rowcolor{mycyan}
        {\bf RemDet-M } & \textbf{28.2} & \bf 46.1 & \bf28.9 & \bf18.2 & \bf41.7 & \bf51.0 & 23.3 & 34.4 \\
        \midrule
        YOLOX-L & 21.3 & 38.4 & 20.4 & 13.0 & 31.0 & 38.1 & 54.2 & 67.7 \\   
        RTMDet-L & 23.7 & 37.4 & 25.5 & 12.5 & 38.7 & 50.4 & 52.3 & 50.4 \\
        YOLOv5-L & 24.5 & 41.0 & 24.9 & 14.2 & 37.1 & 50.7 & 46.2 & 34.0 \\
        YOLOv7-L & 25.4 & 43.2 & 25.2 & 15.6 & 36.7 & 45.8 & 37.2 & 33.0 \\
        YOLOv6-v3.0-L & 26.0 & 42.4 & 26.4 & 15.6 & 38.7 & 48.6 & 59.6 & 47.1 \\
        YOLOv8-L & 27.9 & 45.7 & 28.4 & 17.6 & 41.4 & 52.4 & 43.6 & 82.6 \\
        \rowcolor{mycyan}
        \bf RemDet-L & \bf29.3 & \bf47.4 & \bf30.3 & \bf18.7 & \bf43.4 & \bf55.8 & 35.3 & 66.7 \\
        \midrule
        YOLOX-X & 21.4 & 38.4 & 20.8 & 13.0 & 31.4 & 36.3 & 99.0 & 141 \\
        YOLOv5-X & 25.8 & 42.7 & 26.5 & 14.9 & 39.3 & 54.2 & 86.3 & 64.4 \\
        YOLOv7-X & 27.7 & 46.8 & 27.7 & 17.6 & 39.8 & 50.9 & 70.9 & 59.4 \\
        YOLOv8-X & 28.8 & 46.6 & 29.6 & 18.5 & 42.5 & 55.8 & 90.3 & 138 \\
        \rowcolor{mycyan}
        \bf RemDet-X & \bf29.9 &\bf 48.3 &\bf 31.0 &\bf 19.5 &\bf 44.1 &\bf 58.6 & 74.1 & 112 \\
		\bottomrule
	\end{tabular}
 }
 \caption {RemDet Compare Real-Time Detectors on VisDrone}
\label{tab:appendix_remdet_compare_general_rt_detectors}
\end{table*}

\begin{algorithm*}[tb]
\caption{Pytorch codes of ConvFFN}
\label{alg:algorithm_convffn}
\begin{lstlisting}[language=Python]
class SimpleConvFFN(nn.Module):
    def __init__(self,c1: int,c2: int,e: float = 1,
                 num_blocks: int = 1, add_identity: bool = True, 
                 conv_cfg: OptConfigType = None,
                 norm_cfg: ConfigType = dict(type='BN', momentum=0.03, eps=0.001),
                 act_cfg: ConfigType = dict(type='SiLU', inplace=True),
                 init_cfg: OptMultiConfig = None) -> None:
        super().__init__()
        self.c = int(c2 * e)
        self.cv1 = ConvModule(c1, self.c,
                              kernel_size=1,
                              stride=1,
                              padding=0,
                              conv_cfg=conv_cfg,
                              norm_cfg=norm_cfg,
                              act_cfg=act_cfg)
        self.cv2 = ConvModule(self.c, c2,
                              kernel_size=1,
                              stride=1,
                              padding=0,
                              conv_cfg=conv_cfg,
                              norm_cfg=norm_cfg,
                              act_cfg=None)
        self.add = add_identity and c1 == c2

    def forward(self, x):
        return x + self.cv2(self.cv1(x)) if self.add else self.cv2(self.cv1(x))

\end{lstlisting}
\end{algorithm*}

\begin{algorithm*}[tb]
\caption{Pytorch codes of Multiplication}
\label{alg:algorithm_multiplication}
\begin{lstlisting}[language=Python]
class MultiplicationFFN(nn.Module):
    def __init__(self, c1: int, c2: int, e: float = 0.5,
                 num_blocks: int = 1, add_identity: bool = True,  # shortcut
                 conv_cfg: OptConfigType = None,
                 norm_cfg: ConfigType = dict(type='BN', momentum=0.03, eps=0.001),
                 act_cfg: ConfigType = dict(type='SiLU', inplace=True),
                 init_cfg: OptMultiConfig = None) -> None:
        super().__init__()
        self.c = int(c2 * (e / 2))
        self.cv1 = ConvModule(c1, self.c * 2,
                              kernel_size=1,
                              stride=1,
                              padding=0,
                              conv_cfg=conv_cfg,
                              norm_cfg=norm_cfg,
                              act_cfg=act_cfg)
        self.cv2 = ConvModule(self.c, c2,
                              kernel_size=1,
                              stride=1,
                              padding=0,
                              conv_cfg=conv_cfg,
                              norm_cfg=norm_cfg,
                              act_cfg=None)
        self.act = nn.GELU()
        self.add = add_identity and c1 == c2

    def forward(self, x):
        y, z = self.cv1(x).split((self.c, self.c), 1)
        y = self.act(y) * self.act(z)
        y = self.cv2(y)
        return y + x if self.add else y
\end{lstlisting}
\end{algorithm*}

\end{document}